\documentclass{article}

\usepackage{arxiv}
\usepackage{booktabs}
\usepackage[utf8]{inputenc} 
\usepackage[T1]{fontenc}    
\usepackage{url}            
\usepackage{booktabs}       
\usepackage{amsfonts}       
\usepackage{nicefrac}       
\usepackage{microtype}      
\usepackage{tabularx}
\usepackage{lipsum}         
\usepackage{graphicx}
\usepackage{doi}
\usepackage{algorithm}
\usepackage[toc, acronym, automake,nonumberlist]{glossaries}
\setkeys{glslink}{hyper=false}
\usepackage{algpseudocode}
\usepackage{url,hyperref,lineno,microtype,subcaption}

\usepackage[capitalise]{cleveref}

\makeglossaries
\newacronym{alc}{ALC}{artificial liver classifier}
\newacronym{ai}{AI}{artificial intelligence}
\newacronym{ml}{ML}{machine learning}
\newacronym{ocr}{OCR}{optical character recognition}
\newacronym{rl}{RL}{reinforcement learning}
\newacronym{ann}{ANN}{artificial neural network}
\newacronym{svm}{SVM}{support vector machine}
\newacronym{woa}{WOA}{whale optimization algorithm}
\newacronym{gwo}{GWO}{grey wolf optimization algorithm}
\newacronym{pso}{PSO}{particle swarm optimization}
\newacronym{cfo}{CFO}{crayfish fish optimization}
\newacronym{rrnn}{RRNN}{recursive recurrent neural network}
\newacronym{rlcfd}{RL-CFD}{Riemann-Liouville conformable fractional derivative}
\newacronym{mlp}{MLP}{multi-layer perceptron}
\newacronym{cdsc}{CDSC}{correlation-based dynamic stopping criterion}
\newacronym{if}{IF}{intuitionistic fuzzy}
\newacronym{cs3wd}{CS3WD}{cost-sensitive three-way decisions}
\newacronym{ldm}{LDM}{large margin distribution machine}
\newacronym{lr}{LR}{logistic regression}
\newacronym{rf}{RF}{random forest}
\newacronym{cart}{CART}{classification and regression tree}
\newacronym{lda}{LDA}{linear discriminant analysis}
\newacronym{fox}{FOX}{FOX optimization algorithm}
\newacronym{xgb}{XGB}{XGBoost}
\newacronym{sdg}{SDG}{stochastic gradient descent}
\newacronym{opium}{OPIUM}{online pseudo-inverse update method}
\newacronym{mr}{MR}{multiple regression}

\newacronym{fdo}{FDO}{fitness dependent optimizer}
\newacronym{leo}{LEO}{lagrange elementary optimization}
\newacronym{ana}{ANA}{ant nesting algorithm}

\title{Artificial Liver Classifier: A New Alternative to Conventional Machine Learning Models}

\date{}

\usepackage{authblk}

\newbox{\orcid}\sbox{\orcid}{\includegraphics[scale=0.06]{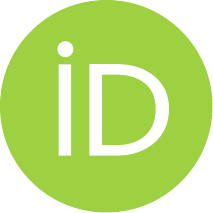}} 
\author[*1]{\href{https://orcid.org/0000-0002-6232-3900}{\usebox{\orcid}\hspace{1mm}Mahmood A. Jumaah}}
\author[1]{\href{https://orcid.org/0000-0002-7216-4149}{\usebox{\orcid}\hspace{1mm}Yossra H. Ali}}
\author[*2]{\href{https://orcid.org/0000-0002-8661-258X}{\usebox{\orcid}\hspace{1mm}Tarik A. Rashid}}
\affil[1]{Department of Computer Science, University of Technology, Baghdad 10066, Iraq.}
\affil[2]{Department of Computer Science and Engineering; AIIC, University of Kurdistan Hewlêr, Erbil 44001, Iraq.}

\affil[*1]{cs.22.27@grad.uotechnology.edu.iq}
\affil[1]{Yossra.H.Ali@uotechnology.edu.iq}
\affil[*2]{tarik.ahmed@ukh.edu.krd}

\begin{document}
\maketitle

\begin{abstract}
	Supervised machine learning classifiers sometimes face challenges related to the performance, accuracy, or overfitting. This paper introduces the Artificial Liver Classifier (ALC), a novel supervised learning model inspired by the human liver's detoxification function. The ALC is characterized by its simplicity, speed, capability to reduce overfitting, and effectiveness in addressing multi-class classification problems through straightforward mathematical operations. To optimize the ALC's parameters, an improved FOX optimization algorithm (IFOX) is employed during training. We evaluate the proposed ALC on five benchmark datasets: Iris Flower, Breast Cancer Wisconsin, Wine, Voice Gender, and MNIST. The results demonstrate competitive performance, with ALC achieving up to 100\% accuracy on the Iris dataset--surpassing logistic regression, multilayer perceptron, and support vector machine--and 99.12\% accuracy on the Breast Cancer dataset, outperforming XGBoost and logistic regression. Across all datasets, ALC consistently shows smaller generalization gaps and lower loss values compared to conventional classifiers. These findings highlight the potential of biologically inspired models to develop efficient machine learning classifiers and open new avenues for innovation in the field.

\end{abstract}
\keywords{Artificial Liver Classifier \and ALC \and Artificial Intelligence \and Classification \and Intelligent Systems \and Machine Learning \and Optimization \and Robotics.}

\section{Introduction}
\label{sec:introduction}
\Gls{ai} has many branches according to the tasks to be performed, with \gls{ml} being one of the most well-known branches that has gained prominence alongside the development of computer science. It focuses on developing systems and algorithms that automatically learn from data without explicit programming~\cite{ai_foundation, ai_future_opportunities,alaa1}. However, two main types of \gls{ml} are categorized according to the problem to be solved: supervised learning and unsupervised learning. Supervised learning relies on having pre-labeled input data (denoted $X$) and the desired output (denoted $y$). This type of learning aims to understand the hidden relationship between inputs and outputs to predict new outcomes based on unseen (new) input data~\cite{supervised_learning_basics, ml_foundation,supervised_learning_applications}. On the other hand, unsupervised learning uses input data that is not pre-labeled (does not contain output y). Instead, an unsupervised learning model is applied to discover patterns and hidden relationships in the data autonomously based on the input data only~\cite{unsupervised_learning_basics, ml_history}. Furthermore, there are other types of \gls{ml} such as \gls{rl}, which interact directly with the problem's environment to build policies that guide decision-making based on rewards and penalties obtained through trial and error~\cite{rl_basics, uot_2, uot_abeer, alaa2}.

In the early stages of \gls{ai}, researchers focused on building systems (with minimal intelligence) capable of performing specific tasks using fixed rules (conditional and logical operations). As the field evolved, scientists realized that intelligent systems needed methods to learn from data, rather than relying on rigid rule-based methods with minimal capabilities~\cite{ai_history,uot_1}. As a result, supervised learning algorithms, specifically classifiers, emerged as tools for learning systems to make predictions or decisions based on the available experiences. However, one of the most preeminent algorithms in supervised learning is \gls{ann}, inspired by the fundamental concept of neurons in the human brain and how they are interconnected~\cite{ml_classification, Abbod2025}. These networks are based on the concept of neurons, which are basic units in the brain that communicate with each other to perform processes such as thinking and learning~\cite{ann_brain_inspired}. The algorithm simulates the functions of brain cells by proposing multiple layers of artificial neurons (an input layer and an output layer). These neurons interact with each other using weights assigned to each connection, and the role of the algorithm is to optimize these weights to minimize the error resulting from interactions with the input data, thereby producing accurate outputs~\cite{foxann}.
Moreover, an older algorithm inspired by mathematics is the \gls{lr}, which aims to find a perfect line that best fits the data points, minimizing the error between actual and predicted labels. These methods were used in statistical analyses before being adopted in \gls{ml}~\cite{linear_regression}. The complexity of linear operations increased, leading to more sophisticated methods, such as \gls{svm}, where the main idea is to create clear boundaries between different data classes by maximizing the margin between them~\cite{improved_svm}. Comprehensively, most of \gls{ml} classifiers have drawn their inspirations from mathematical operations or nature (e.g., simulating the functioning of human brain cells) to create robust systems (classifiers) for solving complex problems. Current \gls{ml} classifiers face multiple challenges related to performance, accuracy or loss, overfitting, and handling data with complex and nonlinear patterns~\cite{ml_challenges, alaa3}.

In this context, this paper proposes a new classifier called \gls{alc}, inspired by the human liver's biological functions. Specifically, it draws on the detoxification function, highlighting its ability to process toxins and convert them into removable forms. Additionally, improvements have been made to \gls{fox}, a state-of-the-art optimization algorithm, to enhance its performance and ensure compatibility with the proposed \gls{alc}. The research aims to bridge the gap in current \gls{ml}'s algorithms by combining the simplicity of mathematical design with solid performance by simulating the detoxification function in the human liver. Furthermore, the proposed classifier aims to improve classification performance by processing data dynamically, simulating the human liver's adaptive ability, enabling its application in fields requiring high-precision solutions and flexibility in dealing with different data patterns. The main challenge lies in transforming the liver's detoxification function into a simplified mathematical model that effectively incorporates properties such as repetition, interaction, and adaptation to the data~\cite{liver_rep_itr_adp}. By comparing the proposed classifier with established \gls{ml} classifiers, the study expects to improve the performance of \gls{ml}, including increased computation speed, better handling of overfitting problems, and avoidance of excessive computational complexity. Additionally, this paper introduces a new concept for drawing inspiration from biological systems, opening up extensive opportunities for researchers to develop mathematical models based on other biological functions of the liver, such as filtering blood or amino acid regulation~\cite{liver_functions}. Moreover, it represents a starting point for interdisciplinary applications combining biology, mathematics, and \gls{ai}, enhancing our understanding of incorporating natural processes into \gls{ml} techniques to create efficient, reliable, and intelligent systems.

The proposed \gls{alc} has been evaluated using a variety of commonly used \gls{ml} datasets, including Wine, Breast Cancer Wisconsin, Iris Flower, MNIST, and Voice Gender~\cite{benchmark_datasets}, which are explained in detail in~\cref{sec:materials}. This diversity in the datasets ensures extensive coverage of different data types, including text, images, and audio, and enables handling binary and multi-class classification problems~\cite{one_class_classification, binary_classification, multiclass_classification}. The purpose of using these datasets is to conduct comprehensive tests to assess the performance of the proposed \gls{alc} and compare it with the established classifiers. The originality and contributions that distinguish this research are as follows:
\begin{enumerate}
	\item	Introducing a new classifier inspired by the liver's biological functions, specifically detoxification, highlighting new possibilities in designing effective classification algorithms based on biological behaviour.

	\item Enhancing the \gls{fox} to improve its performance, address existing limitations, and ensure better compatibility with the proposed \gls{alc}.

	\item Relying on simple mathematical models that simulate the liver's biological interactions, ensuring a balance between design simplicity and high performance.

	\item	Opening new avenues for researchers to draw inspiration from human organ functions, such as the liver, and simulate them in computational ways to contribute innovative solutions for real-world challenges.

	\item Testing the proposed \gls{alc} on diverse datasets demonstrates its effectiveness through experimental results and comparisons with established classifiers.
\end{enumerate}

This paper is structured as follows:~\cref{sec:related works} reviews the literature that has attempted to address classification issues across various data types.~\cref{sec:detoxification} provides an analytical overview of the human liver, focusing on detoxification function and the study's motivation.~\cref{sec:materials and methods} present the used materials and the proposed methodology, including the improvement of classifier design and \gls{fox} training algorithm.~\cref{sec:results,sec:discussion} cover the presentation and analysis of results, including comparisons with previous works. Finally, the study concludes with findings, recommendations, limitations, and future research directions in~\cref{sec:conclusions}.

\section{Related Works}
\label{sec:related works}
This section reviews the standard algorithms used in \gls{ml} classification, with their practical applications across various datasets highlighted~\cite{real_world_applications}. Additionally, recent studies in the field are discussed to identify existing challenges and to shed light on research gaps requiring further attention~\cite{improved_classifications}. Accordingly, the extent to which the proposed classifier can offer practical solutions to these gaps and contribute to the future advancement of the field will be investigated. However, Xiao et al. utilized 12 standard \gls{ml} classifiers on the MNIST dataset, demonstrating its suitability as a benchmark for evaluating the proposed \gls{alc}. Their results identified the Support Vector Classifier (SVC) with a polynomial kernel (C=100) as the best-performing model, achieving an accuracy of 0.978~\cite{r1_fmnist}. This comparable result poses a challenge for the proposed \gls{alc} to surpass. Furthermore, the study~\cite{r2_opium} employed \gls{opium} to classify the MNIST dataset, achieving an accuracy of 0.9590. However, the author noted that these results do not represent cutting-edge methods but rather serve as an instructive baseline and a means of validating the dataset. This makes it feasible to compare the performance of the proposed \gls{alc} against \gls{opium}, as surpassing this baseline would demonstrate an improvement over existing methods. On the other hand, in a comparative study by Cortez et al., three classifiers—\gls{svm}, \gls{mr}, and \gls{ann}—were evaluated on the Wine dataset. The \gls{svm} model demonstrated superior performance, achieving accuracies of 0.8900 for red wine and 0.8600 for white wine, outperforming the other methods with an average accuracy of 0.8790~\cite{r3_wine_svm}. Hence, the findings of Cortez et al. serve as a foundation for further advancements in \gls{ml} applications, providing a basis for evaluating the proposed \gls{alc}.

Another study utilized a \gls{rrnn} on Breast Cancer Wisconsin dataset. The results demonstrated that the proposed model achieved an accuracy of 0.9950~\cite{r2_rrnn}. Despite its outstanding performance, the computational demands of \gls{rrnn} require substantial resources, which may limit their applicability in resource-constrained environments. Moreover, the study~\cite{r6_cs3w} presents a new classification model called CS3W-IFLMC. This model incorporates \gls{if} and \gls{cs3wd} approaches, contributing to improved classification accuracy and reduced costs associated with incorrect decisions. The proposed model has been evaluated using 12 benchmark datasets, demonstrating superior performance compared to \gls{ldm}, F\gls{svm}, and \gls{svm}. However, the study remains limited in scope, as it focuses solely on binary classification tasks and does not extend to multi-class classification problems~\cite{r6_cs3w}. Furthermore, in another study, the researchers examined gender classification (male or female) based on voice data using \gls{mlp}. The findings showed that the \gls{mlp} model outperformed several other methods, including \gls{lr}, \gls{cart}, \gls{rf}, and \gls{svm}. The \gls{mlp} achieved a classification accuracy of 0.9675. This study concluded that the proposed model demonstrates strong discriminative power between genders, which enhances its applicability in auditory data classification tasks~\cite{r7_voice_gender_mlp}.

The reviewed literature, highlights significant advancements in classification models, primarily focusing on improving performance and addressing computational challenges. However, several limitations and research gaps remain. One major issue is the reliance on computationally intensive methods, which can hinder applicability in resource-constrained environments. The absence of practical hyperparameter tuning or reduction mechanisms may also contribute to overfitting and computational inefficiencies. These limitations underscore the need for a new classifier to address such challenges. Hence, the proposed \gls{alc} should emphasize simplicity in design to ensure faster training time with lower cost.

\section{Detoxification in Liver and Motivation}
\label{sec:detoxification}
\begin{figure}[htb]
	\centering
	\includegraphics[width=0.8\textwidth]{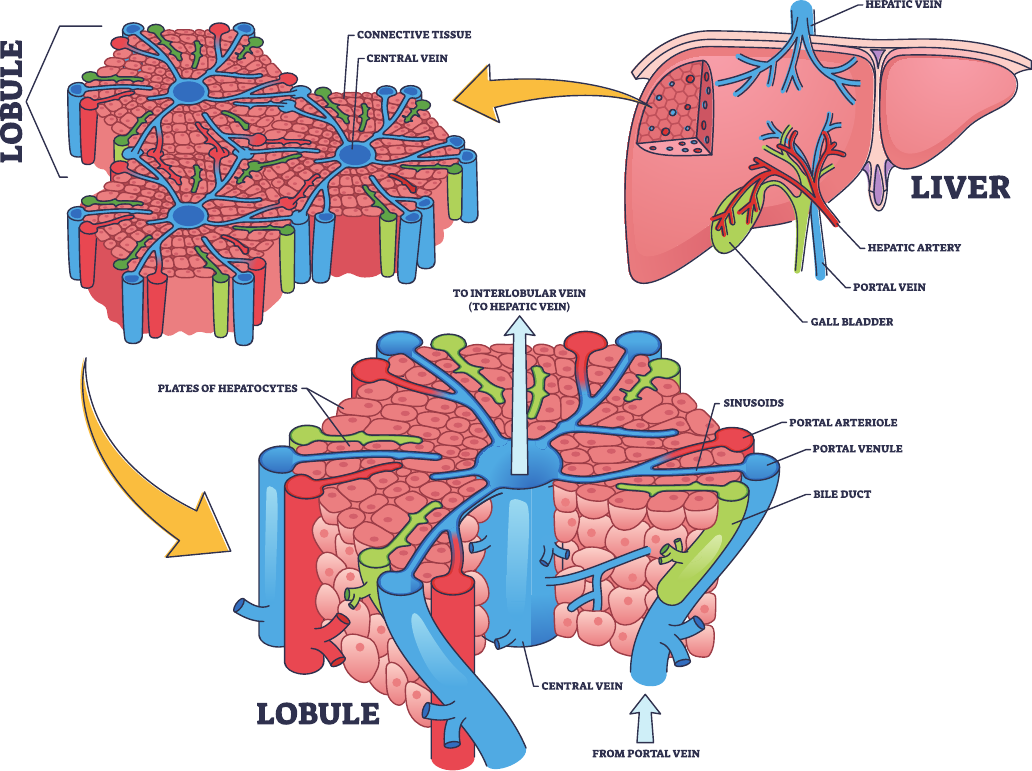}
	\caption{Structural and functional organization of the liver: hepatic lobule and blood flow pathways, concept inspired by~\cite{liver_figure}.}
	\label{fig:liver_anatomy}
\end{figure}

The liver, as illustrated in~\Cref{fig:liver_anatomy}, is the largest internal organ in the human body and is vital in numerous complex physiological processes. It is located in the right upper quadrant of the abdominal cavity and consists of two primary lobes, the right and left, surrounded by a thin membrane known as the hepatic capsule~\cite{human_liver_explain1}. Internally, the liver is composed of microscopic units called hepatic lobules. These hexagonal structures contain hepatic cells organized around a central vein. The lobules are permeated by a network of hepatic sinusoids, which are small channels through which blood flows, facilitating the exchange of oxygen and nutrients between the blood and hepatic cells~\cite{human_liver_explain2}. Furthermore, the liver receives blood from two sources, each contributing different functions. The oxygenated blood enters via the hepatic artery from the aorta, meeting the liver's energy demands. While, the portal vein delivers nutrient-rich and toxin-rich blood from the gastrointestinal tract and spleen~\cite{human_liver_explain3}. The blood from both sources mixes in the hepatic sinusoids, allowing the hepatic cells to perform metabolic and regulatory functions efficiently~\cite{human_liver_explain4}.

However, detoxification is one of the most important liver's functions, which removes toxins from the bloodstream~\cite{detoxification}. Detoxification occurs in two phases. In the phase I, hepatic enzymes known as cytochrome P450 chemically modify toxins through oxidation and reduction reactions, altering their structures to make them more reactive~\cite{cytochrome_P450}. In the phase II, the modified compounds are conjugated with water-soluble molecules such as sulfates or glucuronic acid, making them easier to excrete~\cite{conjugation}. Finally, the toxins are either excreted via bile into the digestive tract or removed from the bloodstream by the kidneys~\cite{excretion_of_nanocarriers}.

The complex biochemical system of the liver has inspired us to develop a new \gls{ml} classifier known as \gls{alc}, modeled after the liver's detoxification mechanisms. The design of the proposed \gls{alc} was guided by an in-depth understanding of the liver's two primary detoxification phases—Cytochrome P450 enzymes and Conjugation pathways—where toxins are transformed into excretable compounds. The proposed \gls{alc} classify feature vectors effectively with minimum training time by simulating these phases using simple \gls{ml} and optimization methods. This innovation marks a significant step forward, demonstrating how biological systems can inspire advanced computational models. It particularly encourages researchers in computer science to explore biological processes for developing intelligent \gls{ml} models.

\section{Materials and Methods}
\label{sec:materials and methods}
This section presents the standard datasets employed for evaluating the proposed \gls{alc} in the conducted experiments. Additionally, the architecture of the proposed \gls{alc} is provided, including mathematical equations, algorithms, and flowcharts. Furthermore, the section elaborates on the \gls{fox}, which serves as the learning algorithm for the proposed \gls{alc}, highlighting its improvements.

\subsection{Materials}
\label{sec:materials}
The following datasets are widely used by \gls{ml} researchers to evaluate their work, making these benchmark datasets suitable for this paper. The MNIST dataset comprises 70,000 grayscale images of handwritten digits (0–9), each of size $28 \times 28$ pixels. It is widely used for multi-class classification tasks due to its diversity and large size~\cite{mnist_dataset}. To utilize the MNIST dataset with the proposed \gls{alc}, each image was preprocessed by flattening it to a vector of 784 dimensions. Each pixel was normalized, with its value transformed to have zero mean and unit variance to ensure consistent scaling. This was then followed by the use of \gls{lda} to project or reduce the data into a low-dimensional space. Hence, \gls{lda} reduces each image to a nine-dimensional feature vector to effectively capture the most discriminative features while also reducing computational requirements~\cite{lda}. Additionally, the Iris dataset, a small-scale collection containing 150 instances across three classes with four features per instance, was included in the proposed \gls{alc} evaluation~\cite{iris_folower_dataset, r1_svm, r3_improved_mlp}. The Breast Cancer Wisconsin dataset, a binary dataset containing 569 samples with 30 features each, was employed to assess the proposed \gls{alc}'s performance on high-dimensional data~\cite{breast_cancer_dataset, r2_rrnn}. Furthermore, the Wine dataset, consisting of 178 samples across three classes with 13 features per instance, was selected for its multi-class nature~\cite{r1_svm, uot_3_woa}. Finally, the Voice Gender dataset was employed to ensure feature diversity. This dataset comprises 3,168 samples, each defined by 21 acoustic features, aimed at distinguishing gender (male or female) by leveraging unique vocal characteristics~\cite{r7_voice_gender_mlp}. These datasets collectively provided a diverse range of classification challenges, enabling a comprehensive evaluation of the proposed \gls{alc}'s performance.
\subsection{Methods}
\label{sec:methods}
This section begins with a detailed introduction to the architecture of the proposed \gls{alc}. Moreover, it delves into the improvements made to the \gls{fox} as a learning algorithm, highlighting its key modifications.

\subsubsection{Artificial Liver Classifier}
\label{sec:artificial liver classifier}
As explained earlier in~\cref{sec:detoxification}, the detoxification process involves the liver's ability to process toxins. Oxygenated blood enters the liver via the hepatic artery, while nutrient-rich blood flows through the portal vein. These sources mix within the hepatic sinusoids, enabling hepatic cells to perform essential functions, including a detoxification function that comprises two phases.

In order to model the detoxification process in the liver, a biologically inspired computational model was chosen to be implemented, where every mathematical operation is merely treated as a simplistic representation of various known mechanisms regarding hepatic detoxification~\cite{detoxification}. The major biochemical steps taken by hepatocytes in the elimination process are considered to include detoxification in two phases: oxidation (Phase I), which is carried out by specialized enzymes and cofactors, and conjugation (Phase II), which is also supported by specific enzymes and cofactors~\cite{conjugation,cytochrome_P450}. In this formulation, these steps are coordinated by being transformed into operations in matrices. A set of molecular input toxins is first linearly overlaid by a cofactor matrix $C$ to model the oxidation action of cytochrome P450, followed by being passed through a nonlinear activation grid to model metabolite selection at a threshold. The second transformation is modeled in the form of conjugation through the interaction of vitamins, and this is subsequently normalized through the use of softmax to represent the classification of detoxified products. Although these operations are not intended to accurately reproduce the actual biochemical kinetics, they are carefully selected so that the structural and functional analogies are preserved—allowing the multi-staged, enzyme-driven, and spatially distributed nature of detoxification to be reproduced within a mathematically consistent and learnable system.

Phase I: toxins are chemically modified to become more reactive. This phase is mathematically simulated by the following equation:

\begin{equation}
	A_{ji} = \frac{1}{n} \sum_{k=1}^{n} (X_{jk} \cdot C_{ki}) + \frac{1}{fp} \sum_{k=1}^{f} \sum_{l=1}^{p} C_{kl}
	\label{eq:p450}
\end{equation}

where $A_{ji}$ is the matrix of reactive toxins, $X$ is the input toxins matrix and $n$ is the number of inputs. The $C$ is initialized randomly within the range $[-1, 1]$ and has dimensions $(f, p)$, where $f$ corresponds to the number of features in the input feature vector, and $p$ is the number of lobules. The term $\frac{1}{fp}\sum_{k=1}^{f} \sum_{l=1}^{p} C_{kl}$ represents the mean of all elements in the cofactor matrix $C$ that used to balance the reaction.

A human liver has a very large number of microscopic functional unit called lobules which are estimated to be of around 100,000~\cite{num_liver_lobules}. In our model the parameter $p$ is an abstraction of a range of choice in lobular diversity, and the columns of the cofactor matrix $C$ implement the various simulated lobular processing units. This structure allows introducing a spatial heterogeneity and an enzymatic variation, which can be witnessed in hepatic tissue, as part of the model. Although the biological premise of $p$ can be that of a number of lobules, direct insertion of a figure like $p = 100, 000$ in computation would be inefficient and also inconvenient because it has a high dimensionality and costs more computations. Thus, we introduce a practical range $f \leq p < 100,000$. This will ensure there is diversity, and that every input feature interacts with a number of simulated lobules ensuring the model is computationally straightforward. The parameter $p$ in this range may be chosen by empirical methods of hyperparameter tuning, a compromise between the richness of the representation, and efficiency. It follows that the range is not arbitrary, but rather biological based on biological modeling under constrains. However, the reactive toxins ($A$) must be activated to enhance their reactivity before progressing to phase II. This activation involves eliminating all negative values, effectively transforming them to zero while retaining only the positive values. This process is mathematically expressed by the following equation:

\begin{equation}
	A' = \mathrm{max}(0, A)
	\label{eq:relu}
\end{equation}
where $A'$ is the activated toxins matrix. However, Equation~\ref{eq:relu} uses ReLU activation to imitate the biological selectivity that occurs in Phase I detoxification in which only reactive (positive) products pass to the next stage. This selection eliminates the non-reactive outputs keeps the computational efficiency and provides the non-linearity necessary in the downstream processing.

Phase II: involves the conjugation of modified compounds from phase I with water-soluble molecules to make them excretable. This phase reduces the toxicity of compounds and facilitates their elimination from the body. this phase can be mathematically modeled using~\Cref{eq:p450}, but with key differences. Instead of toxins, the matrix $A'$ is used as input, representing the modified compounds (activated toxins) generated in phase I. Additionally, a matrix referred to as the vitamin matrix $V$ is employed in place of the cofactor matrix $C$. This matrix is initialized randomly within the range $[-1, 1]$ and has dimensions $(p, n)$.

\begin{equation}
	B_{ji} = \frac{1}{n} \sum_{k=1}^{n} (A_{jk} \cdot V_{ki}) + \frac{1}{pn} \sum_{k=1}^{p} \sum_{l=1}^{n} V_{kl}
	\label{eq:conj}
\end{equation}
where $B_{ji}$ represents the conjugated compounds and $\frac{1}{pn}\sum_{k=1}^{p} \sum_{l=1}^{n} V_{kl}$ represents the mean of all elements in the vitamin matrix $V$.

Lastly, when the reactions in Phase I and Phase II are finished, detoxification is then complete. The outcome is some less dangerous and water-soluble wastes that can be removed by means of bile, urine, stool, etc. The softmax activation function is used as a model of the elimination process: not only does it allow formulating a probabilistic output for each of the classes, but also captures the selective and competitive characteristic of biological excretion~\cite{softmax, softmax_1_tarik, softmax_2}. Many detoxified compounds simultaneously compete to be eliminated in the liver depending on issues such as solvency, the availability of transporters as well as priorities at the cellular level. Softmax reflects this behavior by placing higher probabilities on those compounds that are most dominant, or easily excreted, and thereby simulates the preferential clearance mechanism of the body.

\begin{equation}
	B'_i = \frac{e^{B_i}}{\sum_{j=1}^{n} e^{B_j}}
	\label{eq:softmax}
\end{equation}
where $B'_i$ represents the normalized probability for output class $i$.

\begin{figure}[htb]
	\centering
	\includegraphics[width=0.7\textwidth]{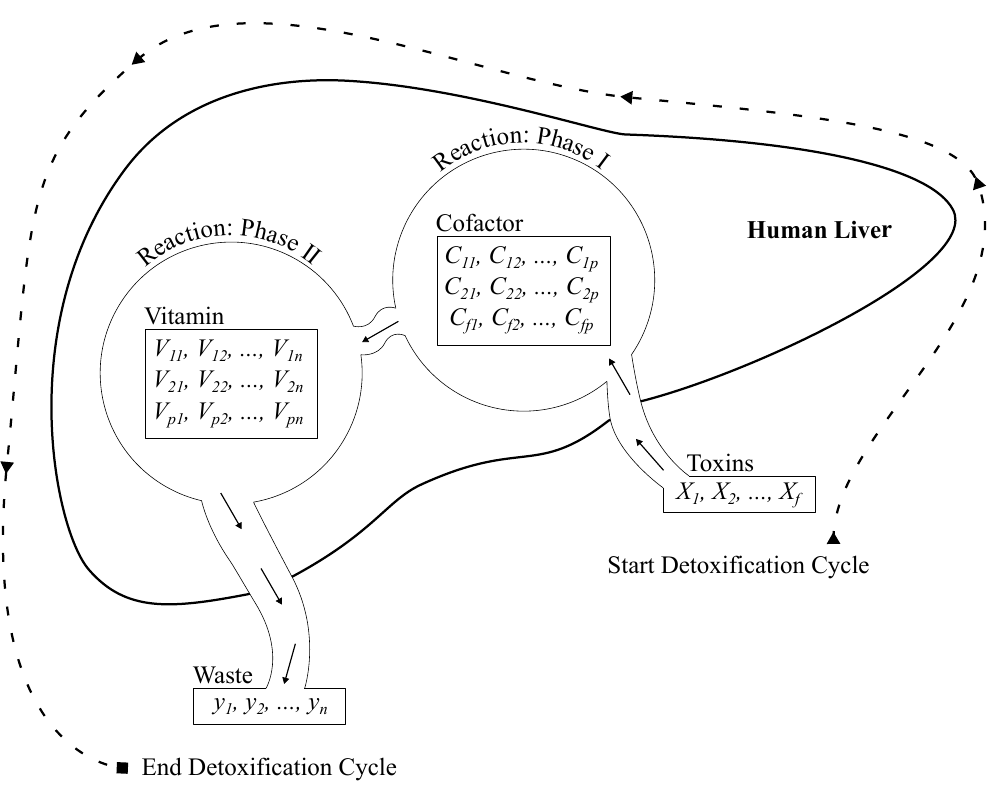}
	\caption{Architecture illustrating Phase I and Phase II reactions simulated by the proposed \gls{alc}, designed to mimic liver detoxification pathways. \label{fig:alc_arch}}
\end{figure}

The ~\Cref{algo:alc} and~\Cref{fig:alc_arch} describes the architecture of the proposed \gls{alc}. First, the cofactor and vitamin matrices are initialized randomly, where these matrices are defined based on the dimensions corresponding to the number of features ($f$), number of lobules ($p$), and number of output classes ($n$). Next, the IFOX, as presented in ~\Cref{algo:fox_new}, is configured, specifying the number of detoxification cycles (maximum number of training epochs) and detoxification power (maximum number of fox agents). The IFOX then optimizes the cofactor and vitamin matrices by minimizing the reaction error (i.e., loss). Finally, the optimized cofactor and vitamin matrices, resulting from this process, are subsequently used together with the toxin inputs (feature vectors) to predict the output classes.

\begin{algorithm}[!h]
	\caption{Artificial Liver Classifier (ALC) \label{algo:alc}}
	\begin{algorithmic}[1]
		\Statex \textbf{Input:} toxins, number of features, number of lobules $p$, number of outputs, detoxification cycles, and detoxification power.
		\Statex \textbf{Output:} predicted classes.
		\State Initialize cofactor matrix $C$ and vitamin matrix $V$ randomly.
		\State Initialize the IFOX training algorithm. \Comment{See~\Cref{algo:fox_new}}
		\State Optimize $C$ and $V$ using IFOX over defined cycles.
		\State Compute reactive toxins. \Comment{using~\Cref{eq:p450}}
		\State Activate reactive toxins. \Comment{Phase I, using~\Cref{eq:relu}}
		\State Perform conjugation to make toxins less harmful. \Comment{Phase II, using~\Cref{eq:conj}}
		\State Normalize outputs to obtain predicted classes. \Comment{Elimination, using~\Cref{eq:softmax}}
		\State \Return predicted classes $\hat{y}$.
	\end{algorithmic}
\end{algorithm}

\begin{figure}[htb]
	\centering
	\includegraphics[width=0.7\textwidth]{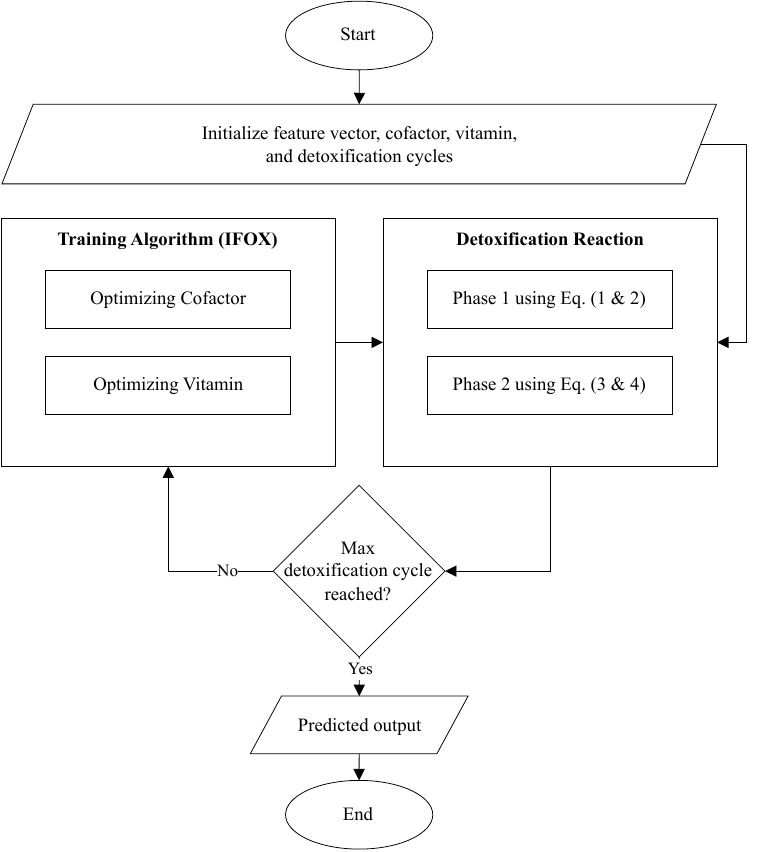}
	\caption{Flowchart of the proposed \gls{alc}. \label{fig:alc_flowchart} }
\end{figure}

Furthermore, the flowchart visualized the proposed \gls{alc} is presented in~\Cref{fig:alc_flowchart}. Additionally, the source code for the implementation of the proposed \gls{alc} can be accessed at the following repository: \url{https://github.com/mwdx93/alc}, which includes the main ALC implementation, training scripts, and example datasets.

\subsubsection{Training Algorithm}
\label{sec:fox optimization algorithm}
The \gls{fox}, developed by Mohammed and Rashid in 2022, mimics the hunting behavior of red foxes by incorporating physics-based principles. These include prey detection based on sound and distance, agent's jumping during the attack governed by gravity, and direction, as well as additional computations such as timing and walking~\cite{fox, foxann}. These features make \gls{fox} a competitive optimization algorithm, outperformed several methods such as \gls{pso} and \gls{fdo}. The \gls{fox} is works as follows: Initially, the ground is covered with snow, requiring the fox agent to search randomly for its prey. During this random search, the fox agent uses the Doppler effect to detect and gradually approach the source of the sound. This process takes time and enables the fox agent to estimate the prey's location by calculating the distance. Once the prey's position is determined, the fox agent computes the required jump to catch it. Additionally, the search process is facilitated through controlled random walks, ensuring the fox agent progresses toward the prey while maintaining an element of randomness. The~\gls{fox} balances exploitation and exploration phases statically, with a 50\% probability for each~\cite{fox_tsa}. Thus, the~\gls{fox} operates as follows:

\begin{enumerate}
	\item Computing the distance $D_i$ of sound travel using the best position and random time:
	      \begin{equation}
		      D_i = \frac{BestPosition}{T_i} \times T_i
		      \label{eq:dist}
	      \end{equation}
	      Where $T_i$ is a random time in $[0, 1]$ and $i$ is the fox agent.

	\item Determining the distance between the fox agent and its prey:
	      \begin{equation}
		      DF_i = 0.5 \times D_i
		      \label{eq:dist_fox}
	      \end{equation}

	\item Computing the jump $J_i$ by multiplying half of the gravity acceleration constant with half squared mean of the time:
	      \begin{equation}
		      J_i = 0.5 \times 9.81 \times 0.5 \times (\sum_{0}^{n}T_i)^2 \label{eq:jump}
	      \end{equation}

	\item Updating the fox agent's position based on a directional equation, either northward $c_1 = 0.18$ or in the opposite direction $c_2 = 0.82$ based on the the jump probability $p$ in $[0, 1]$.

	      \begin{equation}
		      X_{i+1} = DF_i\times J_i\times
		      \begin{cases}
			      c_1, & \text{if } p>0.18 \\
			      c_2, & \text{otherwise}
		      \end{cases}
		      \label{eq:expliot}
	      \end{equation}

	\item The following equation used for exploration:
	      \begin{equation}
		      X_{i+1} = BestPosition \times rand(1, dim) \times Mint \times a \label{eq:explor}
	      \end{equation}
	      where $dim$ is the problem dimension, $Mint$ is the minimum time iteratively updated based on $T_i$, $a$ is an adjustment parameter computed as: $2\times(it - (\frac{1}{Max_{it}}))$, and $it$ is the current iteration.
\end{enumerate}

However, the \gls{fox} has some limitations in its design. These limitations were acknowledged by the  author of \gls{fox}~\cite{fox}, while others have been identified through further analysis. For instance, one notable drawback is its static approach to balancing exploration and exploitation. This paper aims to address these limitations by proposing a new variation of the \gls{fox} called IFOX to make it integrable with the proposed \gls{alc} as a training algorithm to optimize the cofactor and vitamin matrices. For reference, the implementation of the \gls{fox} can be accessed at~\url{https://github.com/hardi-mohammed/fox}.

\begin{algorithm}[!h]
	\caption{IFOX: new variation of FOX optimization algorithm \label{algo:fox_new}}
	\begin{algorithmic}[1]
		\Statex \textbf{Input:} Maximum number of epochs $epochs$, maximum number of fox agents $max_{fa}$
		\Statex \textbf{Output:} $BestX$ and $BestFitness$

		\State Initialize the fox agents population $X_{fa} \quad (fa=1,2,3,...,max_{fa})$
		\State Initialize $BestX, BestFitness$
		\While{$it < epochs$}
		\ForAll {$fa \in \mathcal FAs$}
		\State $f \leftarrow \text{Fitness}(X_{fa})$
		\If {$f < BestFitness$}
		\State $BestFitness \leftarrow f$
		\State $BestX \leftarrow X_{fa}$
		\EndIf
		\EndFor

		\State $\alpha_{min} \leftarrow \frac{1}{2 \times epochs}$
		\State $\alpha \leftarrow \alpha_{min} + (1 - \alpha_{min}) \times (1 - it / epochs)$
		\State $t \leftarrow 0.5 \times \mu(\text{rand}(0, 1, \text{size}(BestX)))$
		\State $Jump \leftarrow 4.905 \times t^2$
		\ForAll {$fa \in \mathcal FAs$}
		\State $\beta \leftarrow \text{rand}(-\alpha, \alpha, \text{size}(BestX))$
		\If {$\text{rand}(0,1) < \alpha$}
		\State $X_{fa} \leftarrow BestX + \beta \times \alpha$
		\Else
		\State $X_{fa} \leftarrow 0.5 \times BestX \times \frac{\beta \times \alpha}{Jump}$
		\EndIf
		\EndFor

		\State $it \leftarrow it + 1$
		\EndWhile

	\end{algorithmic}
\end{algorithm}

The IFOX, as visualized in ~\Cref{algo:fox_new}, incorporates several improvements over the \gls{fox}. First, it transforms the balance between exploitation and exploration into a dynamic process using the $\epsilon$-greedy method, rather than a static approach~\cite{epsilon_greedy, uot_alaa}. This dynamic adjustment is controlled by the parameter $\alpha$, which decreases progressively as the optimization process iterate. Second, the computation of distances is eliminated in favor of directly using the best position, facilitated by the parameter $\beta$, derived from $\alpha$. This modification simplifies the \gls{fox} by removing Equations~\ref{eq:dist} and \ref{eq:dist_fox}, and simplifying~\Cref{eq:expliot} by eliminating the probability parameter $p$ and the directional variables ($c_1$ and $c_2$). Third, in~\Cref{eq:explor}, the variables $a$ and $Mint$ are excluded.

\section{Results}
\label{sec:results}
This section presents the performance results of the proposed \gls{alc} on multiple benchmark datasets, as described in~\cref{sec:materials}. The experimental parameter settings were configured for each dataset as follows: 500 detoxification cycles, a detoxification power of 10, and dataset-specific numbers of lobules. Specifically, the number of lobules was set to 10 for Iris Flower and  Breast Cancer Wisconsin, 15 for Wine and Voice Gender, and 50 for MNIST. The choice of these values was done through a systematic empirical search, where a predetermined set of possible values has been considered, a validation based approach used. Different values of $p$ were searched over each dataset and the value that provided the maximum average classification accuracy on a held-out validation split was selected. This strategy makes sure that the hyperparameters being chosen are tuned in a reproducible and performance-based fashion.
In order to have an efficient and fair assessment on the performance of the models, cross-validation was used on the basis of $k$. In particular, each of the datasets has been split into $k$ equal size folds (10-folds were used in our experiments), and the test has been repeated on every fold. Mean results of each of the runs were used to compute final performance metrics. To facilitate later comparison and analysis, additional classifiers, including \gls{mlp}, \gls{svm}, \gls{lr}, and \gls{xgb}, were executed on the same datasets. However, all experiments were conducted on an MSI GL63 8RD laptop equipped with an Intel® Core™ i7-8750H × 12 processor and 32 GB of memory. This consistent setup ensured a robust evaluation of the proposed \gls{alc} alongside the other classifiers under the same conditions.

\subsection{Performance Metrics}
\label{sec:performance_metrics}
To evaluate the performance of the proposed \gls{alc}, several metrics were employed, including log loss (cross-entropy loss), accuracy, precision, recall, F1-score, and training time. Initially, Log loss ~\Cref{eq:loss} quantifies the divergence between predicted probabilities and actual labels, where lower values indicate better predictive performance~\cite{log_loss}. The accuracy ~\Cref{eq:accuracy} measures the proportion of correctly classified instances, serving as a straightforward indicator of overall correctness. Moreover, precision ~\Cref{eq:precison} evaluates the proportion of true positives among all positive predictions, emphasizing the model's ability to reduce false positives. In contrast, recall ~\Cref{eq:recal} focuses on the proportion of true positives among all actual positive instances, highlighting the importance of minimizing false negatives. Furthermore, the F1-score ~\Cref{eq:f1score}, as the harmonic mean of precision and recall, provides a balanced assessment when class distributions are imbalanced~\cite{evaluation_metrics}. Moreover, the overfitting gap defined as the difference between training and validation accuracy, provides insights into generalization. A smaller value indicate better generalization, while a larger value indicates overfitting, where the model excels on the training set but struggles with unseen data. Finally, the training time reflects the duration required to train the model, offering insight into its computational efficiency.

\begin{equation}
	\text{Log Loss} = -\frac{1}{n} \sum_{i=1}^{n} \left( y_i \log(\hat{y}_i) + (1 - y_i) \log(1 - \hat{y}_i) \right)
	\label{eq:loss}
\end{equation}
\begin{equation}
	\text{Accuracy} = \frac{TP + TN}{TP + FP + TN + FN}
	\label{eq:accuracy}
\end{equation}
\begin{equation}
	\text{Precision} = \frac{TP}{TP + FP}
	\label{eq:precison}
\end{equation}
\begin{equation}
	\text{Recall} = \frac{TP}{TP + FN}
	\label{eq:recal}
\end{equation}
\begin{equation}
	\text{F1-Score} = 2 \times \left( \frac{\text{Precision} \times \text{Recall}}{\text{Precision} + \text{Recall}} \right)
	\label{eq:f1score}
\end{equation}

where TP, TN, FP, and FN represent the true positive, true negative, false positive, and false negative counts, respectively. Additionally, $y$ denotes the actual labels, while $\hat{y} $ represents the predicted labels.

\subsection{Convergence result of the Training algorithm IFOX}
\begin{figure}[!h]
	\includegraphics[width=\textwidth]{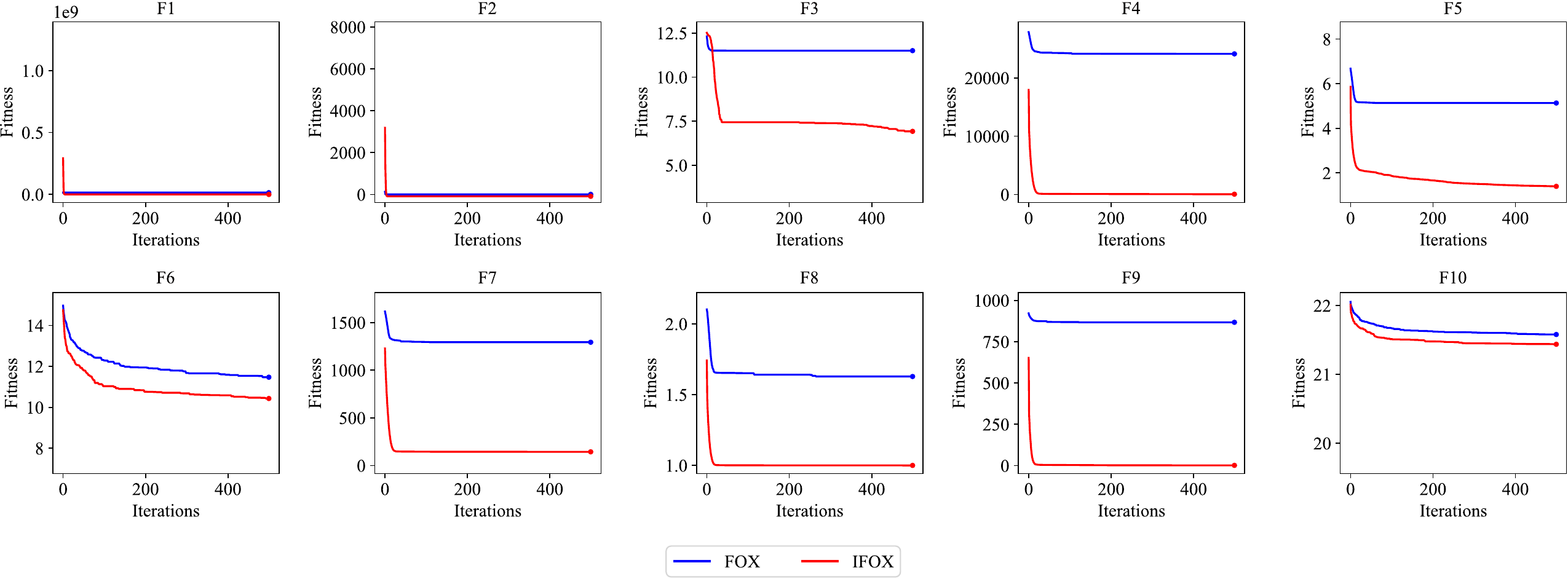}
	\caption{Convergence performance curve of the \gls{fox} (blue) and IFOX (red) on the CEC2019 benchmark test functions. Lower fitness values indicate better convergence performance. \label{fig:new_fox_results} }
\end{figure}

\Cref{tab:fox_ifox_convergence} provides the empirical result of the convergence behavior of IFOX based on selected benchmark functions of the CEC2019 suite. Unlike the original FOX, IFOX invariably reaches lower average objective values in most of the test functions, with vastly less variance and greater stability, especially on high-dimensional and multimodal functions like F3 through F7. Furthermore, four recent and competitive optimizers were added to the test in order to prove the effectiveness of IFOX over its predecessor. They consist of GOOSE, \gls{ana}, \gls{leo}, and \gls{fdo}, which have shown rather good performance in the literature \cite{goose,ana,leo,fdo}. The relative performance assures that IFOX attains better convergence characteristics and final solution quality most of the time. Moreover, the gain noticed in performance increase is explained based on adding adaptive inertia control and better search dynamics in IFOX. It is a design that puts an emphasis on more extensive exploration during the initial phases and more targeted exploitation as optimization goes on. Additionally, convergence curves in Figure~\ref{fig:new_fox_results} demonstrate that IFOX converges more rapidly compared to the original FOX and, in addition, obtains better final values. This can be seen by the fact that the fitness trajectory flattened very fast as compared to other methods. Although formal theoretical demonstration of convergence is outside the scope of this paper, the similar results in independent runs (30 runs) give good evidence of the soundness and reliability of IFOX in varied and different optimization landscapes.

\begin{table}[!h]
	\centering
\caption{Comparison of best objective values obtained by the proposed IFOX algorithm across selected CEC2019 benchmark functions over 30 independent runs. The results are compared with those of FOX and several recent optimization algorithms. Lower objective values indicate better performance.}
	\label{tab:fox_ifox_convergence}
	\begin{tabularx}{\linewidth}{lXXXXXXXXXX}
		\hline
		\textbf{Function/Optimizer} & F1            & F2      & F3            & F4      & F5   & F6            & F7            & F8            & F9            & F10   \\
		\hline
		\textbf{GOOSE}    & 1.8E12      & 6.8E3 & 13.70         & 1.60E3 & 6.09 & 4.79          & 274.35        & 5.57          & 3.81          & 20.98 \\
		\textbf{ANA}      & -             & 4.00    & 13.70         & 38.50   & 1.22 & -             & 116.59        & 5.47          & 2.00          & 2.71  \\
		\textbf{LEO}      & 7.3E9         & 17.47   & 12.70         & 69.86   & 2.76 & 3.01          & 195.56        & 5.06          & 3.26          & 20.01 \\
		\textbf{FDO}      & 4585.27       & 4.00    & 13.7          & 34.08   & 2.13 & 12.13         & 120.40        & 6.1           & 2.01          & 2.71  \\
		\textbf{FOX}      & 1.00          & 4.72    & 9.88          & 147.21  & 5.13 & 298.10        & 1.017         & 1.38          & 1.41          & 21.49 \\
		\textbf{IFOX}     & 1.00 & 4.61    & 2.42 & 35.80   & 1.88 & 1.00 & 1.00 & 1.24 & 1.33 & 20.99 \\
		\hline
	\end{tabularx}
\end{table}

In order to compare performance on the whole, each optimization algorithm was ranked by the number of functions with the best objective value. The overall and the average ranking of all functions is given in \Cref{tab:ifo_rank_summary}. IFOX presented a minimum total (18) and average (1.8) rank and maintained a better performance as compared to the rest of the optimization algorithms. Due to IFOX's superior convergence and stability, it was chosen as the training algorithm in this study.
\begin{table}[!h]
    \centering
    \caption{Function-wise ranks, total rank, and average rank for each optimization algorithm across the 10 selected CEC2019 benchmark functions. Lower ranks indicate better performance.}
    \label{tab:ifo_rank_summary}
    \begin{tabularx}{\linewidth}{l*{10}{X}cc}
        \hline
        \textbf{Function/Optimizer} & \textbf{F1} & \textbf{F2} & \textbf{F3} & \textbf{F4} & \textbf{F5} & \textbf{F6} & \textbf{F7} & \textbf{F8} & \textbf{F9} & \textbf{F10} & \textbf{Total Rank} & \textbf{Avg. Rank} \\
        \hline
        \textbf{IFOX}  & 1 & 3 & 1 & 2 & 2 & 1 & 1 & 1 & 1 & 5 & 18 & 1.8 \\
        \textbf{FDO}   & 3 & 1 & 4 & 1 & 3 & 4 & 4 & 6 & 4 & 1 & 31 & 3.1 \\
        \textbf{ANA}   & 6 & 1 & 4 & 3 & 1 & 6 & 3 & 4 & 3 & 1 & 32 & 3.2 \\
        \textbf{FOX}   & 1 & 4 & 2 & 5 & 5 & 5 & 2 & 2 & 2 & 6 & 34 & 3.4 \\
        \textbf{LEO}   & 4 & 5 & 3 & 4 & 4 & 2 & 5 & 3 & 5 & 3 & 38 & 3.8 \\
        \textbf{GOOSE} & 6 & 6 & 4 & 6 & 6 & 3 & 6 & 5 & 6 & 4 & 52 & 5.2 \\
        \hline
    \end{tabularx}
\end{table}

\subsection{Experimental results of ALC}
\label{sec:experimental_results}
The performance results of the proposed \gls{alc} are presents through this subsection, summarized in the figures and tables. Additionally, comparisons with other classifiers, including \gls{mlp}, \gls{svm}, \gls{lr}, and \gls{xgb}, have been conducted on the five datasets described in~\cref{sec:materials}.

\begin{figure}[!h]
    \centering
    \begin{subfigure}[b]{0.48\textwidth}
        \centering
        \includegraphics[width=\linewidth]{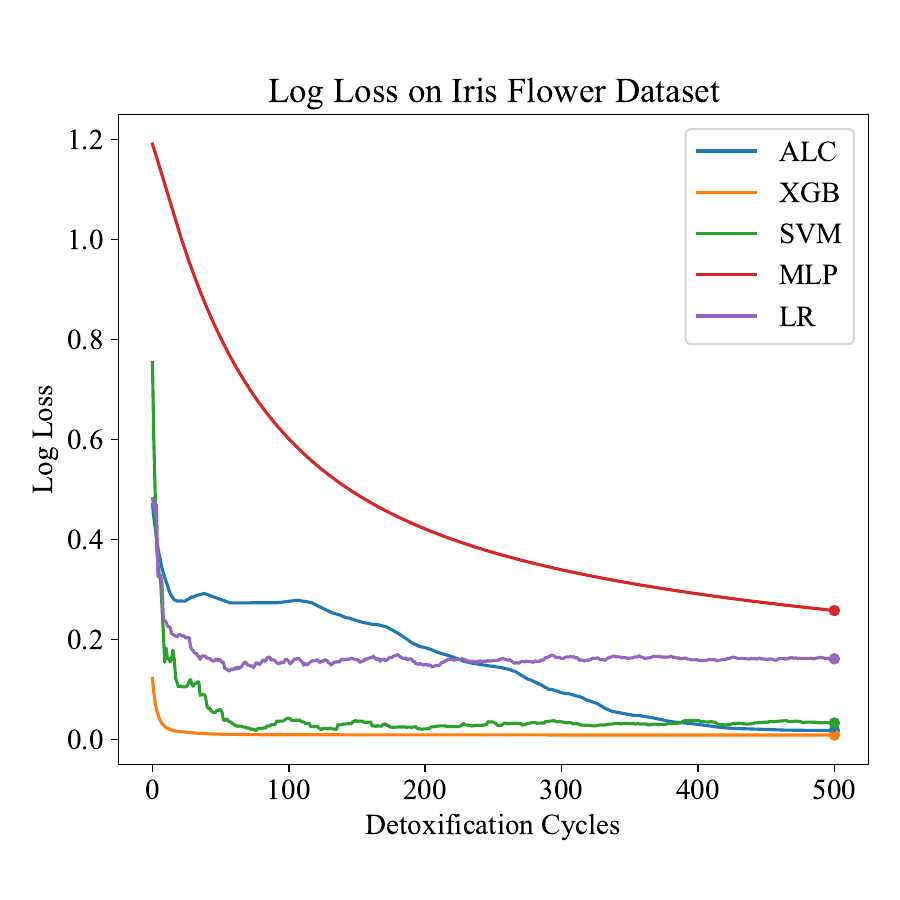}
        \caption{Log loss on the Iris dataset}
        \label{fig:iris_loss}
    \end{subfigure}
    \hfill
    \begin{subfigure}[b]{0.48\textwidth}
        \centering
        \includegraphics[width=\linewidth]{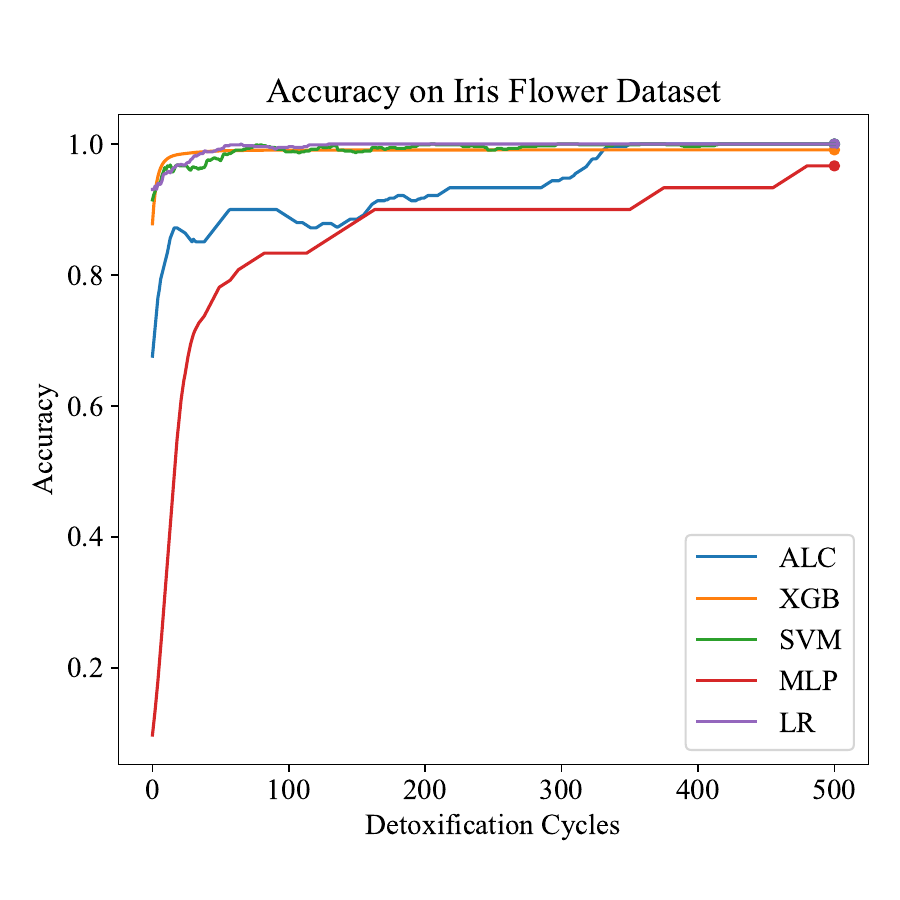}
        \caption{Accuracy on the Iris dataset}
        \label{fig:iris_accuracy}
    \end{subfigure}
    \caption{Performance results comparison of the proposed \gls{alc} (blue) with other classifiers on the validation set of the Iris dataset. Subfigure \textbf{(a)} shows the log loss values, and subfigure \textbf{(b)} shows accuracy.}
    \label{fig:iris_results}
\end{figure}

Table~\ref{tab:iris_results} presents the performance results of the proposed \gls{alc} and other classifiers on the Iris Flower dataset. Additionally, Figures~\ref{fig:iris_loss} and~\ref{fig:iris_accuracy} show the loss and accuracy, respectively, across the validation folds. The proposed \gls{alc} achieved 100\% accuracy with a loss of 0.0169, an overfitting gap of $-0.0231\%$, and a training time of 2.12 seconds. The \gls{xgb} also achieved 100\% accuracy with a loss of 0.0085, an overfitting gap of $-0.0144\%$, and a training time of 0.91 seconds. Similarly, the \gls{svm} reached 100\% accuracy with a loss of 0.0704, an overfitting gap of $-0.0384\%$, and a training time of 4.42 seconds. The \gls{mlp} attained 96.67\% accuracy with a loss of 0.2417, an overfitting gap of $-0.0714\%$, and a training time of 4.18 seconds. Lastly, the \gls{lr} reached 100\% accuracy with a loss of 0.0543, an overfitting gap of $-0.0415\%$, and a training time of 4.31 seconds.
\begin{table}[!h]
	\caption{Cross-validation performance of the proposed \gls{alc} and other classifiers on the Iris Flower dataset (mean over 10-folds).}
	\begin{tabularx}{\linewidth}{XXXXXX}
		\hline
		\textbf{Metric}      & \textbf{\gls{alc}} & \textbf{XGB} & \textbf{SVM} & \textbf{MLP} & \textbf{LR} \\
		\hline
		\textbf{Loss}        & 0.0169             & 0.0085       & 0.0691       & 0.2417       & 0.0543      \\
		\textbf{Accuracy}    & 1.0000             & 1.0000       & 1.0000       & 0.9667       & 1.0000      \\
		\textbf{Precision}   & 1.0000             & 1.0000       & 1.0000       & 0.9694       & 1.0000      \\
		\textbf{Recall}      & 1.0000             & 1.0000       & 1.0000       & 0.9667       & 1.0000      \\
		\textbf{F1-Score}    & 1.0000             & 1.0000       & 1.0000       & 0.9664       & 1.0000      \\
		\textbf{Overfitting} & -0.0231\%          & -0.0144\%    & -0.0384\%    & -0.0689\%    & -0.0415\%   \\
		\textbf{Time (sec.)} & 2.12               & 0.91         & 4.42         & 4.18         & 4.31        \\
		\hline
	\end{tabularx}
	\label{tab:iris_results}
\end{table}

\begin{figure}[!ht]
    \begin{subfigure}[b]{0.48\textwidth}
		\centering
		\includegraphics[width=\linewidth]{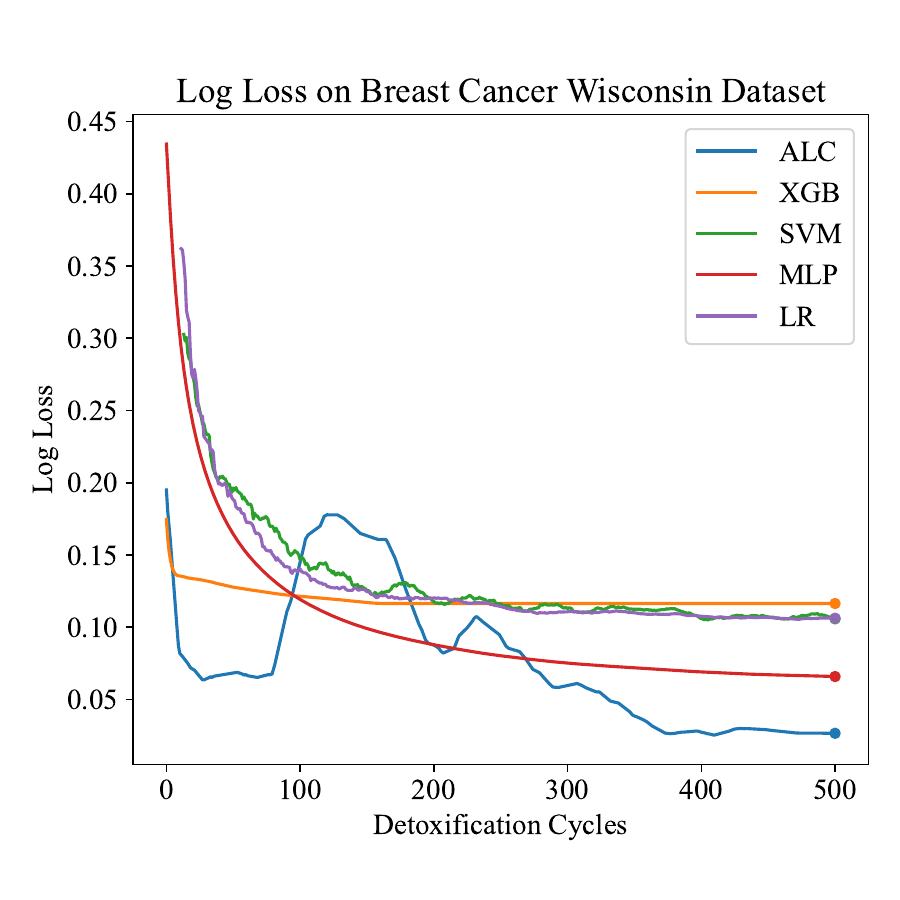}
		\caption{Log loss on the Breast Cancer Wisconsin dataset}
		\label{fig:Breast_loss}
		 \end{subfigure}
	\hfill
    \begin{subfigure}[b]{0.48\textwidth}
		\centering
		\includegraphics[width=\linewidth]{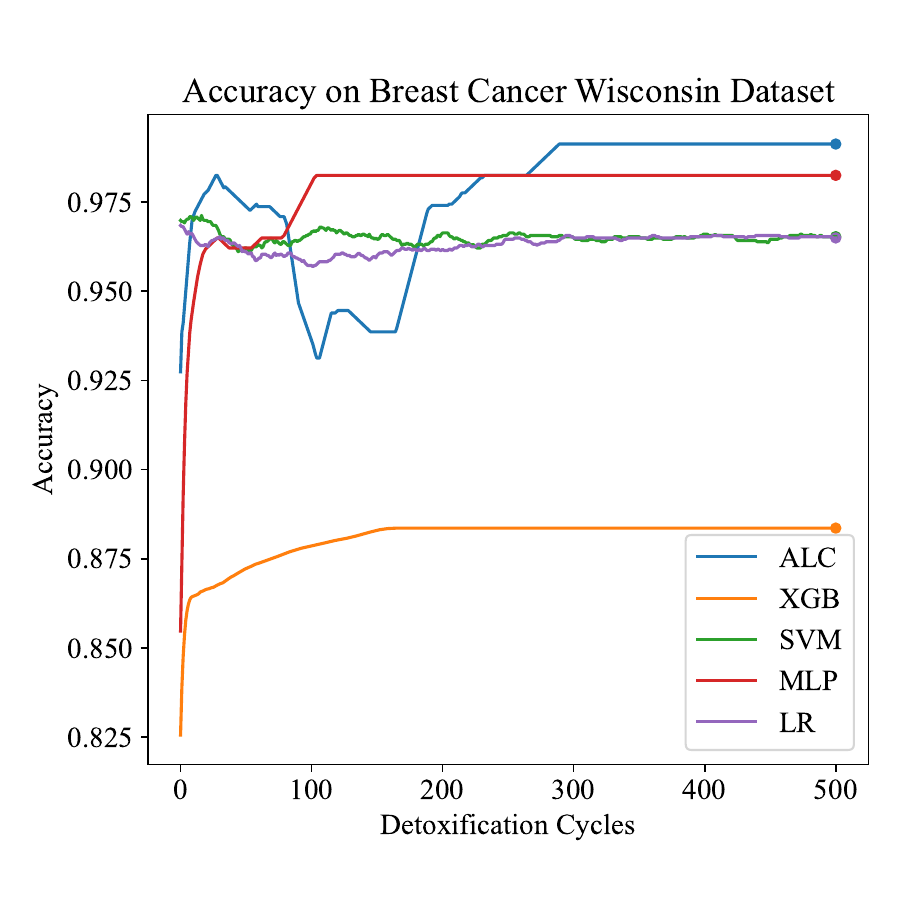}
		\caption{Accuracy on the Breast Cancer Wisconsin dataset}
		\label{fig:Breast_accuracy}
		 \end{subfigure}
	\caption{Performance results comparison of the proposed \gls{alc} (blue) with other classifiers on the validation set of the Breast Cancer Wisconsin dataset. Subfigure \textbf{(a)} shows the log loss values, and subfigure \textbf{(b)} shows accuracy.}
\end{figure}

Table~\ref{tab:breast_cancer_results} presents the performance results of the proposed \gls{alc} and other classifiers on the Breast Cancer Wisconsin dataset. Figures~\ref{fig:Breast_loss} and~\ref{fig:Breast_accuracy} display the loss and accuracy, respectively, on the validation folds. The proposed \gls{alc} achieved 99.12\% accuracy, with a loss of 0.0261 and an overfitting gap of -0.0029\%, with a training time of 3.62 seconds. The \gls{xgb} achieved 88.36\% accuracy, with a loss of 0.1132 and an overfitting gap of 0.1178\%, with a training time of 1.09 seconds. The \gls{svm} reached 98.25\% accuracy, with a loss of 0.1105 and an overfitting gap of 0.0213\%, with a training time of 3.79 seconds. The \gls{mlp} achieved 98.25\% accuracy, with a loss of 0.0682 and an overfitting gap of 0.0086\%, with a training time of 4.53 seconds. Lastly, the \gls{lr} achieved 96.49\% accuracy, with a loss of 0.1319 and an overfitting gap of 0.0237\%, with a training time of 3.81 seconds.
\begin{table}[!h]
	\caption{Cross-validation performance of the proposed \gls{alc} and other classifiers on the Breast Cancer Wisconsin dataset (mean over 10-folds).}
	\begin{tabularx}{\linewidth}{XXXXXX}
		\hline
		\textbf{Metric}      & \textbf{\gls{alc}} & \textbf{XGB} & \textbf{SVM} & \textbf{MLP} & \textbf{LR} \\
		\hline
		\textbf{Loss}        & 0.0261             & 0.1132       & 0.1203       & 0.0682       & 0.1319      \\
		\textbf{Accuracy}    & 0.9932             & 0.8927       & 0.9833       & 0.9833       & 0.9559      \\
		\textbf{Precision}   & 0.9943             & 0.9233       & 0.9833       & 0.9833       & 0.9638      \\
		\textbf{Recall}      & 0.9932             & 0.9233       & 0.9833       & 0.9833       & 0.9559      \\
		\textbf{F1-Score}    & 0.9932             & 0.9260       & 0.9833       & 0.9833       & 0.9631      \\
		\textbf{Overfitting} & -0.0029\%          & 0.1178\%     & 0.0213\%     & 0.0086\%     & 0.0237\%    \\
		\textbf{Time (sec.)} & 3.62               & 1.09         & 3.79         & 4.53         & 3.81        \\
		\hline
	\end{tabularx}
	\label{tab:breast_cancer_results}
\end{table}

\begin{figure}[!ht]
    \centering
    \begin{subfigure}[b]{0.48\textwidth}
        \centering
        \includegraphics[width=\linewidth]{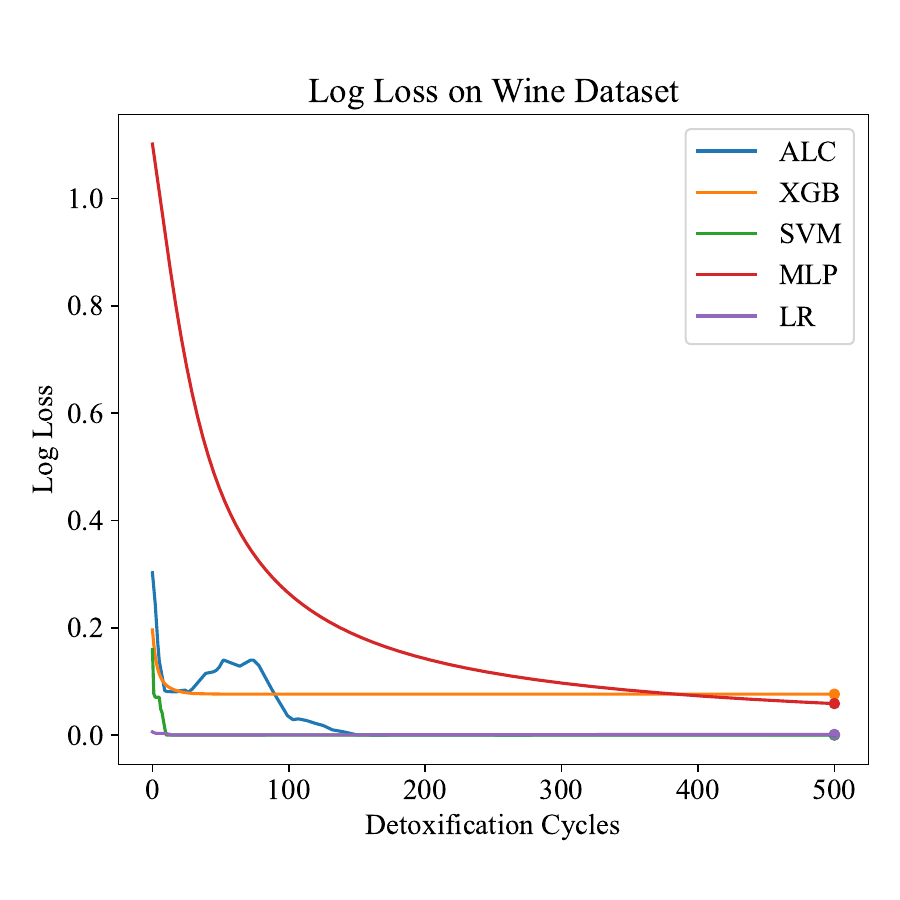}
        \caption{Log loss on the Wine dataset}
        \label{fig:wine_loss}
    \end{subfigure}
    \hfill
    \begin{subfigure}[b]{0.48\textwidth}
        \centering
        \includegraphics[width=\linewidth]{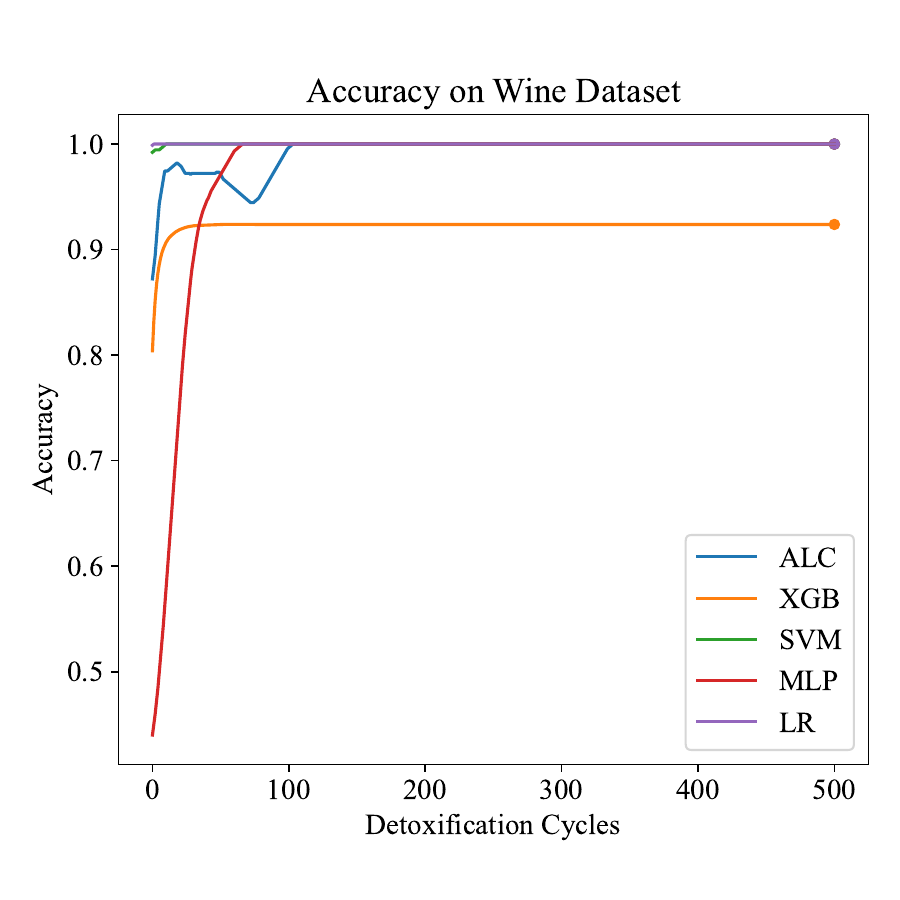}
        \caption{Accuracy on the Wine dataset}
        \label{fig:wine_accuracy}
    \end{subfigure}
    \caption{Performance results comparison of the proposed \gls{alc} (blue) with other classifiers on the validation set of the Wine dataset. Subfigure \textbf{(a)} shows the log loss values, and subfigure \textbf{(b)} shows accuracy.}
    \label{fig:wine_results}
\end{figure}

Table~\ref{tab:wine_results} presents the performance results of the proposed \gls{alc} and other classifiers on the Wine dataset. Figures~\ref{fig:wine_loss} and~\ref{fig:wine_accuracy} display the loss and accuracy, respectively, on the validation folds. The proposed \gls{alc} achieved 100\% accuracy, with a loss of 0.0011 and an overfitting gap of 0.0000\%, with a training time of 2.38 seconds. The \gls{xgb} achieved 92.38\% accuracy, with a loss of 0.0691 and an overfitting gap of 0.0675\%, with a training time of 1.17 seconds. The \gls{svm} achieved 100\% accuracy, with a loss of 0.0001 and an overfitting gap of 0.0000\%, with a training time of 3.89 seconds. The \gls{mlp} achieved 100\% accuracy, with a loss of 0.0568 and an overfitting gap of -0.0068\%, with a training time of 3.89 seconds. Lastly, the \gls{lr} achieved 100\% accuracy, with a loss of 0.0012 and an overfitting gap of 0.0000\%, with a training time of 3.92 seconds.
\begin{table}[!h]
	\caption{Cross-validation performance of the proposed \gls{alc} and other classifiers on the  Wine dataset (mean over 10-folds).}
	\begin{tabularx}{\linewidth}{XXXXXX}
		\hline
		\textbf{Metric}      & \textbf{\gls{alc}} & \textbf{XGB} & \textbf{SVM} & \textbf{MLP} & \textbf{LR} \\
		\hline
		\textbf{Loss}        & 0.0011             & 0.0691       & 0.0001       & 0.0568       & 0.0012      \\
		\textbf{Accuracy}    & 1.0000             & 0.9258       & 1.0000       & 1.0000       & 1.0000      \\
		\textbf{Precision}   & 1.0000             & 0.9534       & 1.0000       & 1.0000       & 1.0000      \\
		\textbf{Recall}      & 1.0000             & 0.9424       & 1.0000       & 1.0000       & 1.0000      \\
		\textbf{F1-Score}    & 1.0000             & 0.9429       & 1.0000       & 1.0000       & 1.0000      \\
		\textbf{Overfitting} & 0.0000\%           & 0.0675\%     & 0.0000\%     & -0.0068\%    & 0.0000\%    \\
		\textbf{Time (sec.)} & 2.38               & 1.17         & 3.89         & 3.89         & 3.92        \\
		\hline
	\end{tabularx}
	\label{tab:wine_results}
\end{table}

\begin{figure}[!ht]
    \centering
    \begin{subfigure}[b]{0.48\textwidth}
        \centering
        \includegraphics[width=\linewidth]{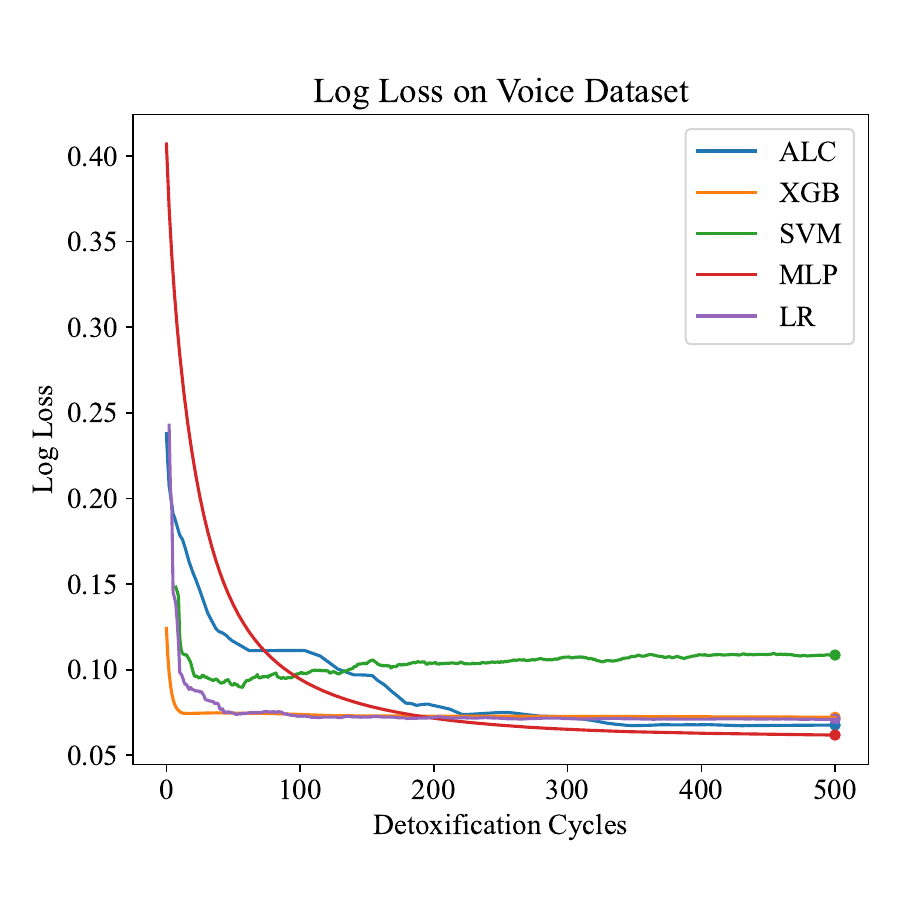}
        \caption{Log loss on the Voice Gender dataset}
        \label{fig:voice_loss}
    \end{subfigure}
    \hfill
    \begin{subfigure}[b]{0.48\textwidth}
        \centering
        \includegraphics[width=\linewidth]{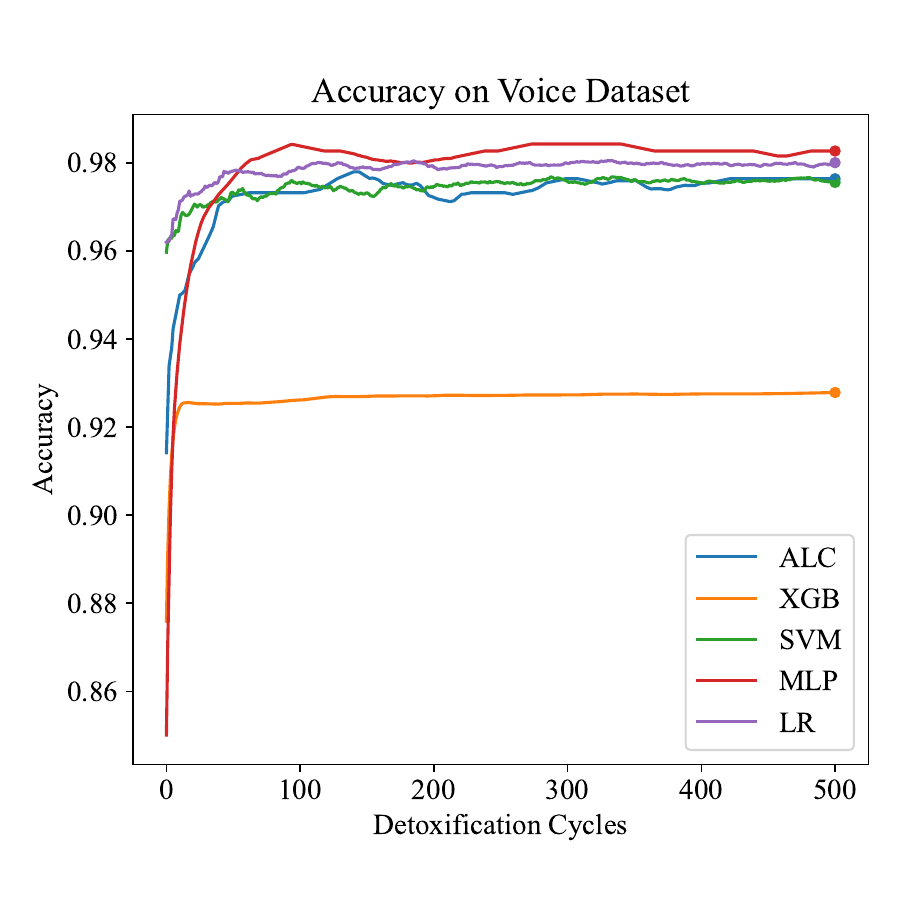}
        \caption{Accuracy on the Voice Gender dataset}
        \label{fig:voice_accuracy}
    \end{subfigure}
    \caption{Performance results comparison of the proposed \gls{alc} (blue) with other classifiers on the validation set of the Voice Gender dataset. Subfigure \textbf{(a)} shows the log loss values, and subfigure \textbf{(b)} shows accuracy.}
    \label{fig:voice_results}
\end{figure}

Table~\ref{tab:voice_results} presents the performance results of the proposed \gls{alc} and other classifiers on the Voice Gender dataset. Figures~\ref{fig:voice_loss} and~\ref{fig:voice_accuracy} display the log loss and accuracy, respectively, on the validation folds. The proposed \gls{alc} achieved 97.63\% accuracy, with a loss of 0.0613 and an overfitting gap of 0.0004\%, with a training time of 3.21 seconds. The \gls{xgb} achieved 92.79\% accuracy, with a loss of 0.0706 and an overfitting gap of 0.0601\%, with a training time of 1.19 seconds. The \gls{svm} achieved 97.32\% accuracy, with a loss of 0.1932 and an overfitting gap of 0.0043\%, with a training time of 4.35 seconds. The \gls{mlp} achieved 98.26\% accuracy, with a loss of 0.0622 and an overfitting gap of -0.0036\%, with a training time of 12.59 seconds. Lastly, the \gls{lr} achieved 98.11\% accuracy, with a loss of 0.0601 and an overfitting gap of -0.0063\%, with a training time of 4.61 seconds.
\begin{table}[!h]
	\caption{Cross-validation performance of the proposed \gls{alc} and other classifiers on the  Voice dataset (mean over 10-folds).}
	\begin{tabularx}{\linewidth}{XXXXXX}
		\hline
		\textbf{Metric}      & \textbf{\gls{alc}} & \textbf{XGB} & \textbf{SVM} & \textbf{MLP} & \textbf{LR} \\
		\hline
		\textbf{Loss}        & 0.0613             & 0.0706       & 0.1932       & 0.0622       & 0.0601      \\
		\textbf{Accuracy}    & 0.9752             & 0.9164       & 0.9750       & 0.9800       & 0.9811      \\
		\textbf{Precision}   & 0.9750             & 0.9325       & 0.9736       & 0.9811       & 0.9811      \\
		\textbf{Recall}      & 0.9752             & 0.9105       & 0.9750       & 0.9800       & 0.9811      \\
		\textbf{F1-Score}    & 0.9752             & 0.9150       & 0.9750       & 0.9811       & 0.9811      \\
		\textbf{Overfitting} & 0.0004\%           & 0.0601\%     & 0.0043\%     & -0.0036\%    & -0.0063\%   \\
		\textbf{Time (sec.)} & 3.21               & 1.19         & 4.35         & 12.59        & 4.61        \\
		\hline
	\end{tabularx}
	\label{tab:voice_results}
\end{table}

\begin{figure}[!ht]
    \centering
    \begin{subfigure}[b]{0.48\textwidth}
        \centering
        \includegraphics[width=\linewidth]{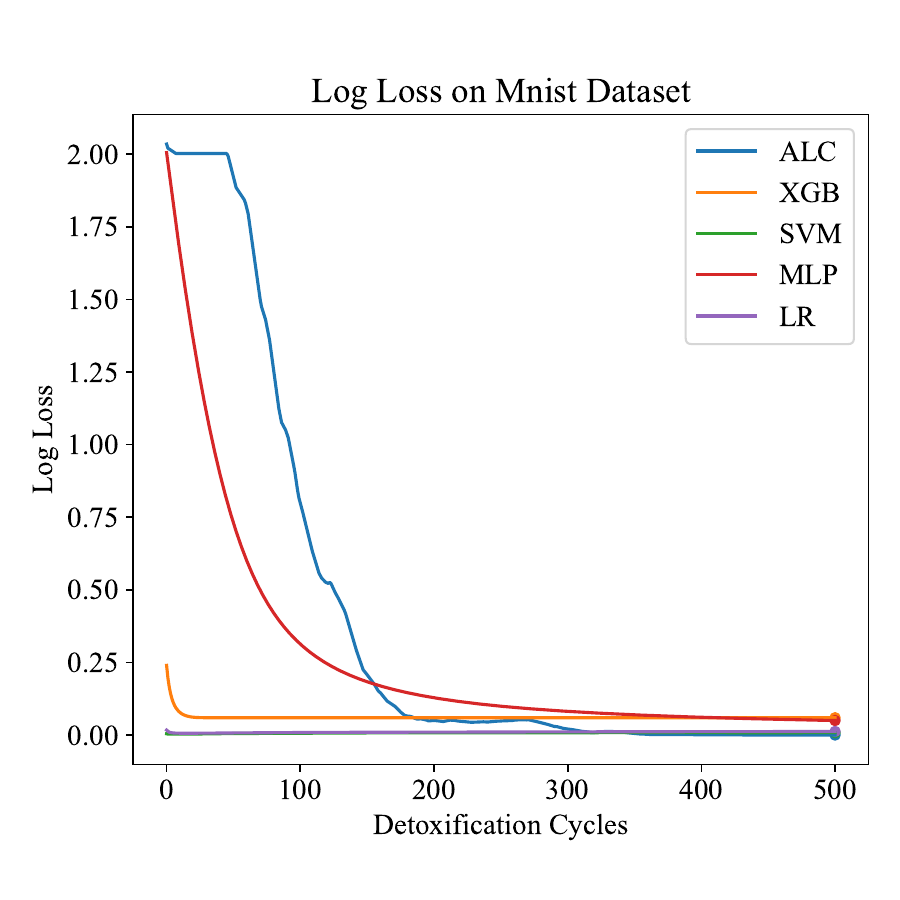}
        \caption{Log loss on the MNIST dataset}
        \label{fig:mnist_loss}
    \end{subfigure}
    \hfill
    \begin{subfigure}[b]{0.48\textwidth}
        \centering
        \includegraphics[width=\linewidth]{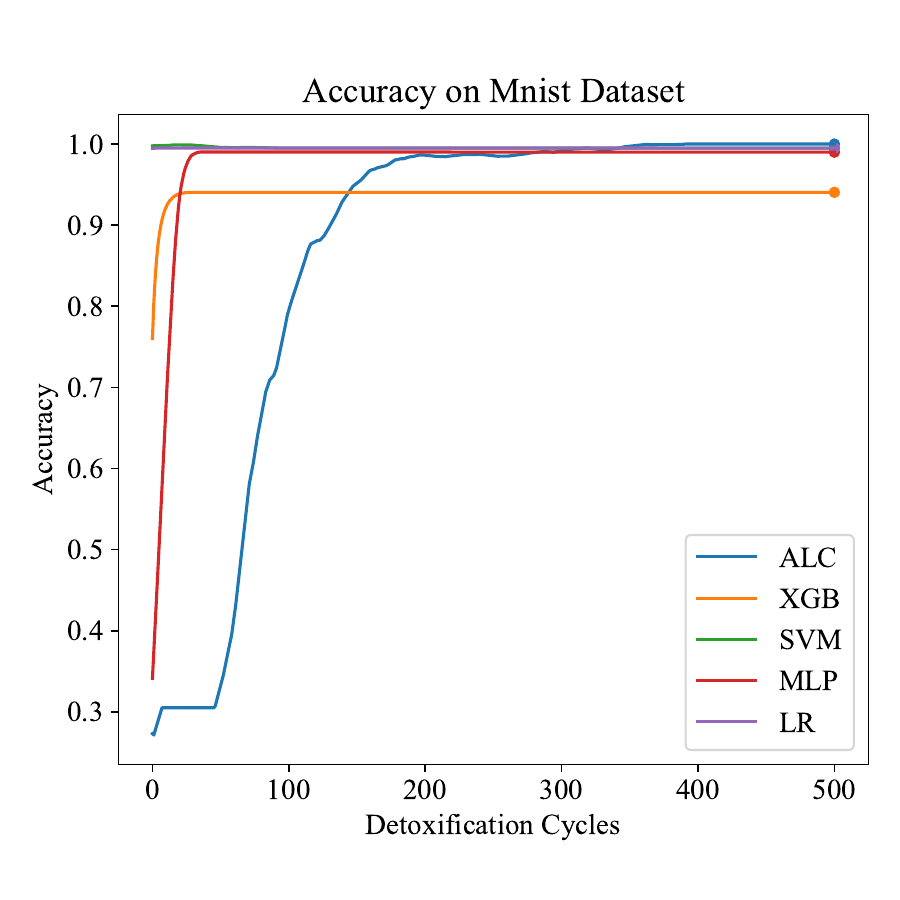}
        \caption{Accuracy on the MNIST dataset}
        \label{fig:mnist_accuracy}
    \end{subfigure}
    \caption{Performance results comparison of the proposed \gls{alc} (blue) with other classifiers on the validation set of the MNIST dataset. Subfigure \textbf{(a)} shows the log loss values, and subfigure \textbf{(b)} shows accuracy.}
    \label{fig:mnist_results}
\end{figure}

Table~\ref{tab:mnist_results} presents the performance results of the proposed \gls{alc} and other classifiers on the MNIST dataset. Figures~\ref{fig:mnist_loss} and~\ref{fig:mnist_accuracy} display the log loss and accuracy, respectively, on the validation set. The proposed \gls{alc} achieved 99.75\% accuracy on the validation set, with a loss of 0.0000 and an overfitting gap of 0.0025\%, with a training time of 6.18 seconds. The \gls{xgb} achieved 94.05\% accuracy, with a loss of 0.0581 and an overfitting gap of 0.0571\%, with a training time of 2.35 seconds. The \gls{svm} achieved 99.50\% accuracy, with a loss of 0.0076 and an overfitting gap of 0.0050\%, with a training time of 5.38 seconds. The \gls{mlp} achieved 99.00\% accuracy, with a loss of 0.0473 and an overfitting gap of 0.0100\%, with a training time of 5.22 seconds. Lastly, the \gls{lr} achieved 99.50\% accuracy, with a loss of 0.0137 and an overfitting gap of 0.0050\%, with a training time of 5.61 seconds.
\begin{table}[!h]
	\caption{Cross-validation performance of the proposed \gls{alc} and other classifiers on the  MNIST dataset (mean over 10-folds).}
	\begin{tabularx}{\linewidth}{XXXXXX}
		\hline
		\textbf{Metric}      & \textbf{\gls{alc}} & \textbf{XGB} & \textbf{SVM} & \textbf{MLP} & \textbf{LR} \\
		\hline
		\textbf{Loss}        & 0.0000             & 0.0581       & 0.0076       & 0.0473       & 0.0137      \\
		\textbf{Accuracy}    & 0.9975             & 0.9421       & 0.9967       & 0.9900       & 0.9967      \\
		\textbf{Precision}   & 0.9970             & 0.9828       & 0.9953       & 0.9906       & 0.9953      \\
		\textbf{Recall}      & 0.9967             & 0.9802       & 0.9967       & 0.9900       & 0.9967      \\
		\textbf{F1-Score}    & 0.9987             & 0.9800       & 0.9967       & 0.9900       & 0.9967      \\
		\textbf{Overfitting} & 0.0025\%           & 0.0571\%     & 0.0050\%     & 0.0100\%     & 0.0050\%    \\
		\textbf{Time (sec.)} & 6.18               & 2.35         & 5.38         & 5.22         & 5.61        \\
		\hline
	\end{tabularx}
	\label{tab:mnist_results}
\end{table}

In summary, the proposed \gls{alc} outperformed or matched other classifiers across the datasets tested, including Iris, Breast Cancer Wisconsin, Wine, Voice Gender, and MNIST. The results demonstrated superior loss, accuracy, precision, recall, and F1-scores, highlighting the reliability and generalization of the proposed \gls{alc} in achieving high classification performance. Furthermore, the proposed \gls{alc} exhibited minimal overfitting and efficient training times compared to other classifiers. However, the next section will provide a detailed analysis and interpretation of these results, and shedding light on limitations and imperfections of the proposed \gls{alc}.

\section{Discussion}
\label{sec:discussion}
The results presented in~\cref{sec:experimental_results}, derived from experiments conducted on the datasets described in~\cref{sec:materials}, highlight the superior performance of the proposed \gls{alc} compared to other classifiers. However, a more in-depth statistical analysis is necessary, particularly of the validation set results, as they are considered more reliable indicators of classifier performance due to being obtained from unseen data. The statistical analysis presented in Table~\ref{tab:results_analysis} compare the performance of the proposed~\gls{alc} with four classifiers—\gls{xgb}, \gls{svm}, \gls{mlp}, and \gls{lr}—across the five datasets described in~\cref{sec:materials}. The analysis focuses on four metrics: loss, accuracy, overfitting gap, and training time, with statistical significance determined using the Wilcoxon signed-rank test at a threshold of $P \text{-value} < 0.05$. The Wilcoxon signed-rank test is used to compare paired samples, particularly when data may not follow a normal distribution. It assesses whether the differences between paired observations are statistically significant~\cite{wilcoxon_signed_rank_test}. Hence, this analysis results provide insights into the strengths of the proposed~\gls{alc} in terms of its generalization, accuracy, and computational efficiency.
\begin{table}[!h]
	\caption{Wilcoxon signed-rank test results comparing classifier pairs on validation set metrics across all datasets.}
	\begin{tabularx}{\linewidth}{llXXXX}
		\hline
		\multicolumn{1}{l}{\textbf{ALC vs.}} & \multicolumn{1}{l}{\textbf{Dataset}} & \multicolumn{4}{c}{\textbf{P-value}}                                                            \\
		\cline{3-6}
		\multicolumn{1}{l}{}                 & \multicolumn{1}{l}{}                 & \textbf{Loss}                        & \textbf{Accuracy} & \textbf{Overfitting} & \textbf{Time} \\
		\hline
		XGB                                  & Iris Flower                          & 0.432                                & 0.872             & 0.481                & 0.493         \\
		SVM                                  & Iris Flower                          & 0.012                                & 0.950             & 0.008                & 0.025         \\
		MLP                                  & Iris Flower                          & 0.006                                & 0.951             & 0.021                & 0.001         \\
		LR                                   & Iris Flower                          & 0.037                                & 0.042             & 0.029                & 0.017         \\

		XGB                                  & Breast Cancer                        & 0.004                                & 0.015             & 0.007                & 0.970         \\
		SVM                                  & Breast Cancer                        & 0.001                                & 0.009             & 0.013                & 0.064         \\
		MLP                                  & Breast Cancer                        & 0.015                                & 0.011             & 0.012                & 0.004         \\
		LR                                   & Breast Cancer                        & 0.000                                & 0.020             & 0.001                & 0.005         \\

		XGB                                  & Wine                                 & 0.022                                & 0.763             & 0.012                & 0.974         \\
		SVM                                  & Wine                                 & 0.893                                & 0.004             & 0.706                & 0.002         \\
		MLP                                  & Wine                                 & 0.005                                & 0.681             & 0.023                & 0.003         \\
		LR                                   & Wine                                 & 0.031                                & 0.822             & 0.748                & 0.002         \\

		XGB                                  & Voice Gender                         & 0.011                                & 0.019             & 0.005                & 0.951         \\
		SVM                                  & Voice Gender                         & 0.000                                & 0.002             & 0.007                & 0.001         \\
		MLP                                  & Voice Gender                         & 0.851                                & 0.804             & 0.029                & 0.002         \\
		LR                                   & Voice Gender                         & 0.019                                & 0.781             & 0.003                & 0.001         \\

		XGB                                  & MNIST                                & 0.001                                & 0.014             & 0.004                & 0.993         \\
		SVM                                  & MNIST                                & 0.003                                & 0.043             & 0.012                & 0.945         \\
		MLP                                  & MNIST                                & 0.001                                & 0.017             & 0.008                & 0.936         \\
		LR                                   & MNIST                                & 0.002                                & 0.041             & 0.009                & 0.898         \\
		\hline
	\end{tabularx}
	\label{tab:results_analysis}
\end{table}
The loss metric, which is a primary indicator of classifier generalizability, demonstrates that the proposed~\gls{alc} outperforms other classifiers in several datasets. Specifically, in the Iris Flower dataset, the proposed~\gls{alc} showed statistically significant improvements in loss compared to \gls{svm} ($P = 0.012$), \gls{mlp} ($P = 0.006$), and \gls{lr} ($P = 0.037$), while its performance was comparable to \gls{xgb} ($P = 0.432$), indicating \gls{xgb} outperforms the proposed \gls{alc}. Similarly, in the Breast Cancer Wisconsin dataset, the proposed~\gls{alc} showed significant improvements over \gls{xgb} ($P = 0.004$), \gls{svm} ($P = 0.001$), \gls{mlp} ($P = 0.015$), and \gls{lr} ($P = 0.000$). In the Wine dataset, the proposed~\gls{alc} demonstrated significant improvements compared to \gls{xgb} ($P = 0.022$), \gls{mlp} ($P = 0.005$), and \gls{lr} ($P = 0.031$), but did not show statistically significant with \gls{svm} ($P = 0.893$). These trends were consistent in more complex datasets like Voice Gender and MNIST, where the proposed~\gls{alc} achieved lower loss values compared to other classifiers in most cases ($P < 0.05$). The findings indicate that the proposed~\gls{alc} offers better generalization across these datasets of varying complexity.

In terms of accuracy, the differences between the proposed~\gls{alc} and other classifiers were generally less pronounced, as reflected by $P$-values exceeding 0.05 in most datasets. Notable exceptions include the Voice Gender dataset, where the proposed~\gls{alc} significantly outperformed \gls{svm} ($P = 0.002$), and the Breast Cancer Wisconsin dataset, where the proposed~\gls{alc} showed an advantage over \gls{xgb} ($P = 0.015$). Furthermore, additional accuracy comparisons were conducted with other models discussed in the related work~\cref{sec:related works}, as presented in Table~\ref{tab:results_comparison_related_work}, demonstrating the superiority of the proposed~\gls{alc}. These results suggest that while accuracy remains an important metric, it may not always effectively differentiate the performance of classifiers, particularly when accuracy levels are already high across classifiers~\cite{accuracy_Confidence}.

\begin{table}[!h]
	\caption{Performance comparison (accuracy metric) of the proposed ALC with models discussed in the related work.}
	\begin{tabularx}{\linewidth}{XXXl}
		\hline
		\textbf{Classifier} & \textbf{Dataset}                   & \textbf{Accuracy} & \textbf{Ref.}               \\
		\hline

		\textbf{ALC}        & \multicolumn{ 1}{l}{Iris Flower}   & \textbf{1.0000}   & Proposed                    \\
		SVM                 & \multicolumn{ 1}{l}{}              & 0.9600            & \cite{r6_cs3w}             \\
		\hline

		ALC                 & \multicolumn{ 1}{l}{Breast Cancer} & 0.9932            & Proposed                    \\
		\textbf{RRNN}       & \multicolumn{ 1}{l}{}              & \textbf{0.9951}   & ~\cite{r2_rrnn}            \\
		\hline

		\textbf{ALC}        & \multicolumn{ 1}{l}{Wine}          & \textbf{1.0000}   & Proposed                    \\
		SVM                 & \multicolumn{ 1}{l}{}              & 0.8790            & \cite{r3_wine_svm}         \\
		MR                  & \multicolumn{ 1}{l}{}              & 0.8645            & \cite{r3_wine_svm}         \\
		ANN                 & \multicolumn{ 1}{l}{}              & 0.8675            & \cite{r3_wine_svm}         \\
		SVM                 & \multicolumn{ 1}{l}{}              & 0.9830            & \cite{r6_cs3w}             \\
		\hline

		\textbf{ALC}        & \multicolumn{ 1}{l}{Voice Gender}  & \textbf{0.9752}   & Proposed                    \\
		MLP                 & \multicolumn{ 1}{l}{}              & 0.9674            & \cite{r7_voice_gender_mlp} \\
		\hline

		\textbf{ALC}        & \multicolumn{ 1}{l}{MNIST}         & \textbf{0.9975}   & Proposed                    \\
		SVC                 & \multicolumn{ 1}{l}{}              & 0.9780            & \cite{r1_fmnist}           \\
		DT                  & \multicolumn{ 1}{l}{}              & 0.8860            & \cite{r1_fmnist}           \\
		KNN                 & \multicolumn{ 1}{l}{}              & 0.9590            & \cite{r1_fmnist}           \\
		MLP                 & \multicolumn{ 1}{l}{}              & 0.9720            & \cite{r1_fmnist}           \\
		OPIUM               & \multicolumn{ 1}{l}{}              & 0.9590            & \cite{r2_opium}            \\

		\hline
	\end{tabularx}
	\label{tab:results_comparison_related_work}
\end{table}

The overfitting gap metric, which evaluates the difference between the performance on training and validation folds, reveals that the proposed~\gls{alc} demonstrates superior generalization. In most datasets, significant improvements were observed, such as in the Breast Cancer Wisconsin and MNIST datasets, where all $P$-values were less than 0.05. In contrast, the overfitting gap in the Wine dataset showed inconsistent patterns, with $P$-values largely exceeding the significance threshold ($P = 0.500$). These results support the ability of the proposed~\gls{alc} to reduce the risk of overfitting. Furthermore, the training time metric is used to measure the speed of classifiers. The statistical results suggest that the proposed~\gls{alc} is competitive and efficient. The training time differed considerably (at least) in datasets that are smaller, including Iris Flower ($P < 0.01$), Breast Cancer Wisconsin ($P = 0.002$), Wine ($P = 0.002$), and voice Gender ($P = 0.001$). But on a bigger and more complicated dataset such as MNIST, the training time of the proposed~\gls{alc} was similar to that of other classifiers ($P > 0.90$). The advocated ~\gls{alc} had no significant differences with \gls{xgb} since it utilized tree-based models, which tend to have short processing times.

\subsection{Computational Complexity and Ablation Analysis}
The complexity of the proposed \gls{alc} is mostly influenced by matrix manipulation and optimization procedure. Suppose that there are n input samples, f features, p lobules, o output classes, and the number of iterations of the optimization (i.e., detoxification cycles) = I. The initial large step is the product of the input toxin matrix $X \in \mathbb{R}^{n \times f}$ by the cofactor matrix $C \in \mathbb{R}^{f \times p}$ in Phase I and takes $\mathcal{O}(nfp)$ time. This is then summed with an element-wise ReLU activation of the resultant matrix whose cost is $\mathcal{O}(np)$. During Phase II, a similar model of conjugation is treated as the second matrix multiplication involving the activated toxin matrix and the vitamin matrix $V$ $\in \mathbb{R}^{npo}$, which would lead to time complexity of $\mathcal{O}(npo)$. This last elimination step runs the softmax on each of the $n$ output vectors, and costs $\mathcal{O}(no)$. The training is based on IFOX that successively optimizes the cofactor and vitamin matrices. Suppose every iteration uses the entire dataset, training will hence have time complexity $\mathcal{O}(I \cdot n \cdot (fp + po))$. So, this term dominates the overall time complexity of the \gls{alc} when training. Moreover, in the space complexity, the model will need $\mathcal{O}(nf)$ storage of the input data, $\mathcal{O}(fp)$ memory to hold the cofactor matrix $C$, $\mathcal{O}(po)$ memory to hold the vitamin matrix $V$, and $\mathcal{O}(np + no)$ intermediate activation and outputs. Thus the overall space complexity is $\mathcal{O}(nf + fp + po + np + no)$. The parameter matrices and batch level intermediate results consume the most memory. Hence, the \gls{alc} has a scalable architecture whose complexity scales linearly with size of input and size of optimization steps and quadratically with size of internal representation (lobules).

The ablation study results were summarized in Table~\ref{tab:alc_ablation}, where the significance of each component of ALC was revealed. The complete model (Phase I + Phase II) represented the optimal result, reaching 99.12 percent accuracy and demonstrating small overfitting. Withdrawing Phase II or replacing Phase I output freedom with a constant value resulted in significant accuracy declines (95.20\% and 91.45\%, respectively), and this fact shows that both steps are needed. The replacement of the cofactor matrix $C$ by random numbers or an identity vitamin matrix also lowered performance indicating the need to learn both matrices. These findings demonstrate that all its components play a significant role in the work of the proposed ALC as a whole.
\begin{table}[!h]
	\centering
	\caption{Ablation study results showing the contribution of Phase I, Phase II, and the respective matrices in the proposed ALC on the Breast Cancer Wilcoxon dataset.}
	\begin{tabularx}{\linewidth}{lXXXXX}
		\hline
		\textbf{Model Variant}        & \textbf{Accuracy} & \textbf{Loss} & \textbf{F1-Score} & \textbf{Overfitting} & \textbf{Time (sec.)} \\
		\hline
		Full ALC (Phase I + Phase II) & 99.12\%           & 0.0261        & 0.9932            & -0.0029\%            & 3.62                 \\
		Phase I only                  & 95.20\%           & 0.0745        & 0.9517            & 0.0121\%             & 2.58                 \\
		Phase II only                 & 91.45\%           & 0.1123        & 0.9140            & 0.0450\%             & 2.89                 \\
		Random cofactor matrix        & 88.36\%           & 0.1341        & 0.8823            & 0.0663\%             & 2.51                 \\
		Identity vitamin matrix       & 93.62\%           & 0.0897        & 0.9312            & 0.0387\%             & 2.73                 \\
		\hline
	\end{tabularx}
	\label{tab:alc_ablation}
\end{table}

\subsection{Failure Case Analysis}
Although the proposed~\gls{alc} can deliver good results irrespective of the data encountered, a few limitations can be associated with it, based on an application during certain situations. ALC does not utilize any mini-batch training mechanism, e.g., stochastic gradient descent~\cite{sdg}, which is normally applied to large-scale learning to minimize computing costs. This consequence can cause longer runtimes in full-batch training working with mass data. There is also slower convergence in the model in that it uses the stochastic IFOX that does not directly optimize training error when applied to cofactor and vitamin matrices. It could influence either the stability or efficiency of convergence. From a model behavior perspective, ALC might fail to perform well on datasets with poor nonlinear structure, noisy or sparse features, or extreme class skew, where the biological metaphor might not find any useful patterns. In addition to that, errors can be propagated and replicated by the sequential dependency between Phase I and Phase II. These constraints point to the directions of further research, such as utilizing mini-batch techniques, improving the IFOX, implementing hybrid optimization schemes, or reorganizing the bio-chemical paradigm to be less rigid and more flexible.

\section{Conclusions}
\label{sec:conclusions}
In conclusion, this paper suggests a novel supervised learning classifier, termed the artificial liver classifier (ALC), inspired by the human liver's detoxification function. The ALC is easy to implement, fast, and capable of reducing overfitting by simulating the detoxification function through straightforward mathematical operations. Furthermore, it introduces an improvement to the FOX optimization algorithm, referred to as IFOX, which is integrated with the ALC as training algorithm to optimize parameters effectively. Furthermore, the ALC was evaluated on five benchmark machine learning datasets: Iris Flower, Breast Cancer Wisconsin, Wine, Voice Gender, and MNIST. The empirical results demonstrated its superior performance compared to support vector machines, multilayer perceptron, logistic regression, XGBoost and other established classifiers. Despite these superiority, the ALC has limitations, such as longer training times on large datasets and slower convergence rates, which could be addressed in future work using methods like mini-batch training or parallel processing. Finally, this paper underscores the potential of biologically inspired models and encourages researchers to simulate natural functions to develop more efficient and powerful machine learning models.

\bibliographystyle{elsarticle-num}
\bibliography{references}

\begin{thebibliography}{10}
\expandafter\ifx\csname url\endcsname\relax
  \def\url#1{\texttt{#1}}\fi
\expandafter\ifx\csname urlprefix\endcsname\relax\def\urlprefix{URL }\fi
\expandafter\ifx\csname href\endcsname\relax
  \def\href#1#2{#2} \def\path#1{#1}\fi

\bibitem{ai_foundation}
A.~Kolides, A.~Nawaz, A.~Rathor, D.~Beeman, M.~Hashmi, S.~Fatima, D.~Berdik, M.~Al-Ayyoub, Y.~Jararweh, \href{http://dx.doi.org/10.1016/j.simpat.2023.102754}{Artificial intelligence foundation and pre-trained models: Fundamentals, applications, opportunities, and social impacts}, Simulation Modelling Practice and Theory 126 (2023) 102754.
\newblock \href {https://doi.org/10.1016/j.simpat.2023.102754} {\path{doi:10.1016/j.simpat.2023.102754}}.
\newline\urlprefix\url{http://dx.doi.org/10.1016/j.simpat.2023.102754}

\bibitem{ai_future_opportunities}
Y.~K. Dwivedi, L.~Hughes, E.~Ismagilova, G.~Aarts, C.~Coombs, T.~Crick, Y.~Duan, R.~Dwivedi, J.~Edwards, A.~Eirug, V.~Galanos, P.~V. Ilavarasan, M.~Janssen, P.~Jones, A.~K. Kar, H.~Kizgin, B.~Kronemann, B.~Lal, B.~Lucini, R.~Medaglia, K.~Le~Meunier-FitzHugh, L.~C. Le~Meunier-FitzHugh, S.~Misra, E.~Mogaji, S.~K. Sharma, J.~B. Singh, V.~Raghavan, R.~Raman, N.~P. Rana, S.~Samothrakis, J.~Spencer, K.~Tamilmani, A.~Tubadji, P.~Walton, M.~D. Williams, \href{http://dx.doi.org/10.1016/j.ijinfomgt.2019.08.002}{Artificial intelligence (ai): Multidisciplinary perspectives on emerging challenges, opportunities, and agenda for research, practice and policy}, International Journal of Information Management 57 (2021) 101994.
\newblock \href {https://doi.org/10.1016/j.ijinfomgt.2019.08.002} {\path{doi:10.1016/j.ijinfomgt.2019.08.002}}.
\newline\urlprefix\url{http://dx.doi.org/10.1016/j.ijinfomgt.2019.08.002}

\bibitem{alaa1}
A.~T. Khudhair, A.~T. Maolood, E.~K. Gbashi, \href{http://dx.doi.org/10.3390/sym16070872}{Symmetry analysis in construction two dynamic lightweight s-boxes based on the 2d tinkerbell map and the 2d duffing map}, Symmetry 16~(7) (2024) 872.
\newblock \href {https://doi.org/10.3390/sym16070872} {\path{doi:10.3390/sym16070872}}.
\newline\urlprefix\url{http://dx.doi.org/10.3390/sym16070872}

\bibitem{supervised_learning_basics}
T.~Jiang, J.~L. Gradus, A.~J. Rosellini, \href{http://dx.doi.org/10.1016/j.beth.2020.05.002}{Supervised machine learning: A brief primer}, Behavior Therapy 51~(5) (2020) 675–687.
\newblock \href {https://doi.org/10.1016/j.beth.2020.05.002} {\path{doi:10.1016/j.beth.2020.05.002}}.
\newline\urlprefix\url{http://dx.doi.org/10.1016/j.beth.2020.05.002}

\bibitem{ml_foundation}
K.~S. Beam, J.~A.~F. Zupancic, \href{http://dx.doi.org/10.1038/s41390-022-02420-1}{Machine learning: remember the fundamentals}, Pediatric Research 93~(2) (2022) 291–292.
\newblock \href {https://doi.org/10.1038/s41390-022-02420-1} {\path{doi:10.1038/s41390-022-02420-1}}.
\newline\urlprefix\url{http://dx.doi.org/10.1038/s41390-022-02420-1}

\bibitem{supervised_learning_applications}
Z.~Zhao, L.~Alzubaidi, J.~Zhang, Y.~Duan, Y.~Gu, \href{http://dx.doi.org/10.1016/j.eswa.2023.122807}{A comparison review of transfer learning and self-supervised learning: Definitions, applications, advantages and limitations}, Expert Systems with Applications 242 (2024) 122807.
\newblock \href {https://doi.org/10.1016/j.eswa.2023.122807} {\path{doi:10.1016/j.eswa.2023.122807}}.
\newline\urlprefix\url{http://dx.doi.org/10.1016/j.eswa.2023.122807}

\bibitem{unsupervised_learning_basics}
D.~S. Watson, \href{http://dx.doi.org/10.1007/s13347-023-00635-6}{On the philosophy of unsupervised learning}, Philosophy \& Technology 36~(2) (Apr. 2023).
\newblock \href {https://doi.org/10.1007/s13347-023-00635-6} {\path{doi:10.1007/s13347-023-00635-6}}.
\newline\urlprefix\url{http://dx.doi.org/10.1007/s13347-023-00635-6}

\bibitem{ml_history}
C.~Molnar, G.~Casalicchio, B.~Bischl, \href{http://dx.doi.org/10.1007/978-3-030-65965-3_28}{Interpretable Machine Learning – A Brief History, State-of-the-Art and Challenges}, Springer International Publishing, 2020, p. 417–431.
\newblock \href {https://doi.org/10.1007/978-3-030-65965-3_28} {\path{doi:10.1007/978-3-030-65965-3_28}}.
\newline\urlprefix\url{http://dx.doi.org/10.1007/978-3-030-65965-3_28}

\bibitem{rl_basics}
Q.~Gao, A.~M. Schweidtmann, \href{http://dx.doi.org/10.1016/j.coche.2024.101012}{Deep reinforcement learning for process design: Review and perspective}, Current Opinion in Chemical Engineering 44 (2024) 101012.
\newblock \href {https://doi.org/10.1016/j.coche.2024.101012} {\path{doi:10.1016/j.coche.2024.101012}}.
\newline\urlprefix\url{http://dx.doi.org/10.1016/j.coche.2024.101012}

\bibitem{uot_2}
H.~Mutar, M.~Jawad, \href{https://ijccce.uotechnology.edu.iq/article\_178569.html}{Analytical study for optimization techniques to prolong wsns life}, IRAQI JOURNAL OF COMPUTERS, COMMUNICATIONS, CONTROL AND SYSTEMS ENGINEERING 23~(2) (2023) 13--23.
\newblock \href {https://doi.org/https://doi.org/10.33103/uot.ijccce.23.2.2} {\path{doi:https://doi.org/10.33103/uot.ijccce.23.2.2}}.
\newline\urlprefix\url{https://ijccce.uotechnology.edu.iq/article\_178569.html}

\bibitem{uot_abeer}
M.~H. Ismael, A.~T. Maolood, \href{http://dx.doi.org/10.1016/j.matpr.2021.05.694}{Withdrawn: Developing modern system in healthcare to detect covid 19 based on internet of things}, Materials Today: Proceedings (Jun. 2021).
\newblock \href {https://doi.org/10.1016/j.matpr.2021.05.694} {\path{doi:10.1016/j.matpr.2021.05.694}}.
\newline\urlprefix\url{http://dx.doi.org/10.1016/j.matpr.2021.05.694}

\bibitem{alaa2}
A.~T. Khudhair, A.~T. Maolood, E.~K. Gbashi, \href{http://dx.doi.org/10.3390/telecom5030044}{Symmetric keys for lightweight encryption algorithms using a pre–trained vgg16 model}, Telecom 5~(3) (2024) 892–906.
\newblock \href {https://doi.org/10.3390/telecom5030044} {\path{doi:10.3390/telecom5030044}}.
\newline\urlprefix\url{http://dx.doi.org/10.3390/telecom5030044}

\bibitem{ai_history}
A.~Grzybowski, K.~Pawlikowska–Łagód, W.~C. Lambert, \href{http://dx.doi.org/10.1016/j.clindermatol.2023.12.016}{A history of artificial intelligence}, Clinics in Dermatology 42~(3) (2024) 221–229.
\newblock \href {https://doi.org/10.1016/j.clindermatol.2023.12.016} {\path{doi:10.1016/j.clindermatol.2023.12.016}}.
\newline\urlprefix\url{http://dx.doi.org/10.1016/j.clindermatol.2023.12.016}

\bibitem{uot_1}
S.~Jabber, S.~Hashem, S.~Jafer, \href{https://ijccce.uotechnology.edu.iq/article\_178584.html}{Analytical and comparative study for optimization problems}, IRAQI JOURNAL OF COMPUTERS, COMMUNICATIONS, CONTROL AND SYSTEMS ENGINEERING 23~(4) (2023) 46--57.
\newblock \href {https://doi.org/https://doi.org/10.33103/uot.ijccce.23.4.5} {\path{doi:https://doi.org/10.33103/uot.ijccce.23.4.5}}.
\newline\urlprefix\url{https://ijccce.uotechnology.edu.iq/article\_178584.html}

\bibitem{ml_classification}
A.~Palanivinayagam, C.~Z. El-Bayeh, R.~Damaševičius, \href{http://dx.doi.org/10.3390/a16050236}{Twenty years of machine-learning-based text classification: A systematic review}, Algorithms 16~(5) (2023) 236.
\newblock \href {https://doi.org/10.3390/a16050236} {\path{doi:10.3390/a16050236}}.
\newline\urlprefix\url{http://dx.doi.org/10.3390/a16050236}

\bibitem{Abbod2025}
A.~A. Abbod, A.~K.~A. Hassan, M.~A. Jumaah, \href{http://dx.doi.org/10.1063/5.0254607}{Analyzing user behavior for targeted commercial advertisements using apriori and k-means algorithms}, in: 4TH INTERNATIONAL CONFERENCE ON INNOVATION IN IOT, ROBOTICS AND AUTOMATION (IIRA 4.0), Vol. 3224, AIP Publishing, 2025, p. 030017.
\newblock \href {https://doi.org/10.1063/5.0254607} {\path{doi:10.1063/5.0254607}}.
\newline\urlprefix\url{http://dx.doi.org/10.1063/5.0254607}

\bibitem{ann_brain_inspired}
S.~Schmidgall, R.~Ziaei, J.~Achterberg, L.~Kirsch, S.~P. Hajiseyedrazi, J.~Eshraghian, \href{http://dx.doi.org/10.1063/5.0186054}{Brain-inspired learning in artificial neural networks: A review}, APL Machine Learning 2~(2) (May 2024).
\newblock \href {https://doi.org/10.1063/5.0186054} {\path{doi:10.1063/5.0186054}}.
\newline\urlprefix\url{http://dx.doi.org/10.1063/5.0186054}

\bibitem{foxann}
M.~A. Jumaah, Y.~H. Ali, T.~A. Rashid, S.~Vimal, \href{http://dx.doi.org/10.70403/3008-1084.1001}{Foxann: A method for boosting neural network performance}, Journal of Soft Computing and Computer Applications 1~(1) (Jun. 2024).
\newblock \href {https://doi.org/10.70403/3008-1084.1001} {\path{doi:10.70403/3008-1084.1001}}.
\newline\urlprefix\url{http://dx.doi.org/10.70403/3008-1084.1001}

\bibitem{linear_regression}
E.~Jumin, N.~Zaini, A.~N. Ahmed, S.~Abdullah, M.~Ismail, M.~Sherif, A.~Sefelnasr, A.~El-Shafie, \href{http://dx.doi.org/10.1080/19942060.2020.1758792}{Machine learning versus linear regression modelling approach for accurate ozone concentrations prediction}, Engineering Applications of Computational Fluid Mechanics 14~(1) (2020) 713–725.
\newblock \href {https://doi.org/10.1080/19942060.2020.1758792} {\path{doi:10.1080/19942060.2020.1758792}}.
\newline\urlprefix\url{http://dx.doi.org/10.1080/19942060.2020.1758792}

\bibitem{improved_svm}
Z.~Quan, L.~Pu, \href{http://dx.doi.org/10.1007/s10639-022-11514-6}{An improved accurate classification method for online education resources based on support vector machine (svm): Algorithm and experiment}, Education and Information Technologies 28~(7) (2022) 8097–8111.
\newblock \href {https://doi.org/10.1007/s10639-022-11514-6} {\path{doi:10.1007/s10639-022-11514-6}}.
\newline\urlprefix\url{http://dx.doi.org/10.1007/s10639-022-11514-6}

\bibitem{ml_challenges}
S.~Tufail, H.~Riggs, M.~Tariq, A.~I. Sarwat, \href{http://dx.doi.org/10.3390/electronics12081789}{Advancements and challenges in machine learning: A comprehensive review of models, libraries, applications, and algorithms}, Electronics 12~(8) (2023) 1789.
\newblock \href {https://doi.org/10.3390/electronics12081789} {\path{doi:10.3390/electronics12081789}}.
\newline\urlprefix\url{http://dx.doi.org/10.3390/electronics12081789}

\bibitem{alaa3}
A.~T. Khudhair, A.~T. Maolood, E.~K. Gbashi, \href{http://dx.doi.org/10.70403/3008-1084.1003}{A novel approach to generate dynamic s-box for lightweight cryptography based on the 3d hindmarsh rose model}, Journal of Soft Computing and Computer Applications 1~(1) (Jun. 2024).
\newblock \href {https://doi.org/10.70403/3008-1084.1003} {\path{doi:10.70403/3008-1084.1003}}.
\newline\urlprefix\url{http://dx.doi.org/10.70403/3008-1084.1003}

\bibitem{liver_rep_itr_adp}
Y.~Tan, K.~An, J.~Su, \href{http://dx.doi.org/10.1016/j.cbpc.2024.109925}{Review: Mechanism of herbivores synergistically metabolizing toxic plants through liver and intestinal microbiota}, Comparative Biochemistry and Physiology Part C: Toxicology \& Pharmacology 281 (2024) 109925.
\newblock \href {https://doi.org/10.1016/j.cbpc.2024.109925} {\path{doi:10.1016/j.cbpc.2024.109925}}.
\newline\urlprefix\url{http://dx.doi.org/10.1016/j.cbpc.2024.109925}

\bibitem{liver_functions}
H.~Ishibashi, M.~Nakamura, A.~Komori, K.~Migita, S.~Shimoda, \href{http://dx.doi.org/10.1007/s00281-009-0155-6}{Liver architecture, cell function, and disease}, Seminars in Immunopathology 31~(3) (2009) 399–409.
\newblock \href {https://doi.org/10.1007/s00281-009-0155-6} {\path{doi:10.1007/s00281-009-0155-6}}.
\newline\urlprefix\url{http://dx.doi.org/10.1007/s00281-009-0155-6}

\bibitem{benchmark_datasets}
F.~Hoffmann, T.~Bertram, R.~Mikut, M.~Reischl, O.~Nelles, \href{http://dx.doi.org/10.1002/widm.1318}{Benchmarking in classification and regression}, WIREs Data Mining and Knowledge Discovery 9~(5) (Jun. 2019).
\newblock \href {https://doi.org/10.1002/widm.1318} {\path{doi:10.1002/widm.1318}}.
\newline\urlprefix\url{http://dx.doi.org/10.1002/widm.1318}

\bibitem{one_class_classification}
N.~Seliya, A.~Abdollah~Zadeh, T.~M. Khoshgoftaar, \href{http://dx.doi.org/10.1186/s40537-021-00514-x}{A literature review on one-class classification and its potential applications in big data}, Journal of Big Data 8~(1) (Sep. 2021).
\newblock \href {https://doi.org/10.1186/s40537-021-00514-x} {\path{doi:10.1186/s40537-021-00514-x}}.
\newline\urlprefix\url{http://dx.doi.org/10.1186/s40537-021-00514-x}

\bibitem{binary_classification}
P.~N, M.~G, \href{http://dx.doi.org/10.1109/ICCCNT56998.2023.10307442}{A review: Binary classification and hybrid segmentation of brain stroke using transfer learning-based approach}, in: 2023 14th International Conference on Computing Communication and Networking Technologies (ICCCNT), IEEE, 2023, p. 1–6.
\newblock \href {https://doi.org/10.1109/icccnt56998.2023.10307442} {\path{doi:10.1109/icccnt56998.2023.10307442}}.
\newline\urlprefix\url{http://dx.doi.org/10.1109/ICCCNT56998.2023.10307442}

\bibitem{multiclass_classification}
B.~Sidumo, E.~Sonono, I.~Takaidza, \href{http://dx.doi.org/10.1016/j.ecoinf.2022.101822}{An approach to multi-class imbalanced problem in ecology using machine learning}, Ecological Informatics 71 (2022) 101822.
\newblock \href {https://doi.org/10.1016/j.ecoinf.2022.101822} {\path{doi:10.1016/j.ecoinf.2022.101822}}.
\newline\urlprefix\url{http://dx.doi.org/10.1016/j.ecoinf.2022.101822}

\bibitem{real_world_applications}
I.~H. Sarker, \href{http://dx.doi.org/10.1007/s42979-021-00592-x}{Machine learning: Algorithms, real-world applications and research directions}, SN Computer Science 2~(3) (Mar. 2021).
\newblock \href {https://doi.org/10.1007/s42979-021-00592-x} {\path{doi:10.1007/s42979-021-00592-x}}.
\newline\urlprefix\url{http://dx.doi.org/10.1007/s42979-021-00592-x}

\bibitem{improved_classifications}
B.~F. Azevedo, A.~M. A.~C. Rocha, A.~I. Pereira, \href{http://dx.doi.org/10.1007/s10994-023-06467-x}{Hybrid approaches to optimization and machine learning methods: a systematic literature review}, Machine Learning 113~(7) (2024) 4055–4097.
\newblock \href {https://doi.org/10.1007/s10994-023-06467-x} {\path{doi:10.1007/s10994-023-06467-x}}.
\newline\urlprefix\url{http://dx.doi.org/10.1007/s10994-023-06467-x}

\bibitem{r1_fmnist}
H.~Xiao, K.~Rasul, R.~Vollgraf, \href{https://arxiv.org/abs/1708.07747}{Fashion-mnist: a novel image dataset for benchmarking machine learning algorithms} (2017).
\newblock \href {https://doi.org/10.48550/ARXIV.1708.07747} {\path{doi:10.48550/ARXIV.1708.07747}}.
\newline\urlprefix\url{https://arxiv.org/abs/1708.07747}

\bibitem{r2_opium}
G.~Cohen, S.~Afshar, J.~Tapson, A.~van Schaik, \href{https://arxiv.org/abs/1702.05373}{Emnist: an extension of mnist to handwritten letters} (2017).
\newblock \href {https://doi.org/10.48550/ARXIV.1702.05373} {\path{doi:10.48550/ARXIV.1702.05373}}.
\newline\urlprefix\url{https://arxiv.org/abs/1702.05373}

\bibitem{r3_wine_svm}
P.~Cortez, A.~Cerdeira, F.~Almeida, T.~Matos, J.~Reis, \href{http://dx.doi.org/10.1016/j.dss.2009.05.016}{Modeling wine preferences by data mining from physicochemical properties}, Decision Support Systems 47~(4) (2009) 547–553.
\newblock \href {https://doi.org/10.1016/j.dss.2009.05.016} {\path{doi:10.1016/j.dss.2009.05.016}}.
\newline\urlprefix\url{http://dx.doi.org/10.1016/j.dss.2009.05.016}

\bibitem{r2_rrnn}
V.~Rajeswari, K.~Sakthi~Priya, \href{http://dx.doi.org/10.1016/j.bspc.2024.106810}{Ontological modeling with recursive recurrent neural network and crayfish optimization for reliable breast cancer prediction}, Biomedical Signal Processing and Control 99 (2025) 106810.
\newblock \href {https://doi.org/10.1016/j.bspc.2024.106810} {\path{doi:10.1016/j.bspc.2024.106810}}.
\newline\urlprefix\url{http://dx.doi.org/10.1016/j.bspc.2024.106810}

\bibitem{r6_cs3w}
S.~Fan, H.~Li, C.~Guo, D.~Liu, L.~Zhang, \href{http://dx.doi.org/10.1016/j.ins.2024.120726}{A novel cost-sensitive three-way intuitionistic fuzzy large margin classifier}, Information Sciences 674 (2024) 120726.
\newblock \href {https://doi.org/10.1016/j.ins.2024.120726} {\path{doi:10.1016/j.ins.2024.120726}}.
\newline\urlprefix\url{http://dx.doi.org/10.1016/j.ins.2024.120726}

\bibitem{r7_voice_gender_mlp}
M.~Buyukyilmaz, A.~O. Cibikdiken, \href{http://dx.doi.org/10.2991/msota-16.2016.90}{Voice gender recognition using deep learning}, in: Proceedings of 2016 International Conference on Modeling, Simulation and Optimization Technologies and Applications (MSOTA2016), msota-16, Atlantis Press, 2016.
\newblock \href {https://doi.org/10.2991/msota-16.2016.90} {\path{doi:10.2991/msota-16.2016.90}}.
\newline\urlprefix\url{http://dx.doi.org/10.2991/msota-16.2016.90}

\bibitem{human_liver_explain1}
E.~Moradi, S.~Jalili-Firoozinezhad, M.~Solati-Hashjin, \href{http://dx.doi.org/10.1016/j.actbio.2020.08.041}{Microfluidic organ-on-a-chip models of human liver tissue}, Acta Biomaterialia 116 (2020) 67–83.
\newblock \href {https://doi.org/10.1016/j.actbio.2020.08.041} {\path{doi:10.1016/j.actbio.2020.08.041}}.
\newline\urlprefix\url{http://dx.doi.org/10.1016/j.actbio.2020.08.041}

\bibitem{human_liver_explain2}
C.~C. Kennedy, E.~E. Brown, N.~O. Abutaleb, G.~A. Truskey, \href{http://dx.doi.org/10.3389/fcvm.2021.625016}{Development and application of endothelial cells derived from pluripotent stem cells in microphysiological systems models}, Frontiers in Cardiovascular Medicine 8 (Feb. 2021).
\newblock \href {https://doi.org/10.3389/fcvm.2021.625016} {\path{doi:10.3389/fcvm.2021.625016}}.
\newline\urlprefix\url{http://dx.doi.org/10.3389/fcvm.2021.625016}

\bibitem{human_liver_explain3}
A.~Schlegel, H.~Mergental, C.~Fondevila, R.~J. Porte, P.~J. Friend, P.~Dutkowski, \href{http://dx.doi.org/10.1016/j.jhep.2023.02.009}{Machine perfusion of the liver and bioengineering}, Journal of Hepatology 78~(6) (2023) 1181–1198.
\newblock \href {https://doi.org/10.1016/j.jhep.2023.02.009} {\path{doi:10.1016/j.jhep.2023.02.009}}.
\newline\urlprefix\url{http://dx.doi.org/10.1016/j.jhep.2023.02.009}

\bibitem{human_liver_explain4}
A.~Gibert-Ramos, D.~Sanfeliu-Redondo, P.~Aristu-Zabalza, A.~Martínez-Alcocer, J.~Gracia-Sancho, S.~Guixé-Muntet, A.~Fernández-Iglesias, \href{http://dx.doi.org/10.3390/cancers13225719}{The hepatic sinusoid in chronic liver disease: The optimal milieu for cancer}, Cancers 13~(22) (2021) 5719.
\newblock \href {https://doi.org/10.3390/cancers13225719} {\path{doi:10.3390/cancers13225719}}.
\newline\urlprefix\url{http://dx.doi.org/10.3390/cancers13225719}

\bibitem{detoxification}
G.~Donati, A.~Angeletti, L.~Gasperoni, F.~Piscaglia, A.~L. Croci~Chiocchini, A.~Scrivo, T.~Natali, I.~Ullo, C.~Guglielmo, P.~Simoni, R.~Mancini, L.~Bolondi, G.~La~Manna, \href{http://dx.doi.org/10.1007/s40620-020-00799-w}{Detoxification of bilirubin and bile acids with intermittent coupled plasmafiltration and adsorption in liver failure (hercole study)}, Journal of Nephrology 34~(1) (2020) 77–88.
\newblock \href {https://doi.org/10.1007/s40620-020-00799-w} {\path{doi:10.1007/s40620-020-00799-w}}.
\newline\urlprefix\url{http://dx.doi.org/10.1007/s40620-020-00799-w}

\bibitem{cytochrome_P450}
F.~P. Guengerich, \href{http://dx.doi.org/10.1007/s43188-020-00056-z}{A history of the roles of cytochrome p450 enzymes in the toxicity of drugs}, Toxicological Research 37~(1) (2020) 1–23.
\newblock \href {https://doi.org/10.1007/s43188-020-00056-z} {\path{doi:10.1007/s43188-020-00056-z}}.
\newline\urlprefix\url{http://dx.doi.org/10.1007/s43188-020-00056-z}

\bibitem{conjugation}
H.~Sun, K.~S. Schanze, \href{http://dx.doi.org/10.1021/acsami.2c02475}{Functionalization of water-soluble conjugated polymers for bioapplications}, ACS Applied Materials \&; Interfaces 14~(18) (2022) 20506–20519.
\newblock \href {https://doi.org/10.1021/acsami.2c02475} {\path{doi:10.1021/acsami.2c02475}}.
\newline\urlprefix\url{http://dx.doi.org/10.1021/acsami.2c02475}

\bibitem{excretion_of_nanocarriers}
A.~Zhang, K.~Meng, Y.~Liu, Y.~Pan, W.~Qu, D.~Chen, S.~Xie, \href{http://dx.doi.org/10.1016/j.cis.2020.102261}{Absorption, distribution, metabolism, and excretion of nanocarriers in vivo and their influences}, Advances in Colloid and Interface Science 284 (2020) 102261.
\newblock \href {https://doi.org/10.1016/j.cis.2020.102261} {\path{doi:10.1016/j.cis.2020.102261}}.
\newline\urlprefix\url{http://dx.doi.org/10.1016/j.cis.2020.102261}

\bibitem{liver_figure}
M.~R. Nikmaneshi, B.~Firoozabadi, L.~L. Munn, \href{http://dx.doi.org/10.1002/bit.27451}{A mechanobiological mathematical model of liver metabolism}, Biotechnology and Bioengineering 117~(9) (2020) 2861–2874.
\newblock \href {https://doi.org/10.1002/bit.27451} {\path{doi:10.1002/bit.27451}}.
\newline\urlprefix\url{http://dx.doi.org/10.1002/bit.27451}

\bibitem{mnist_dataset}
E.~R. G, S.~M, A.~R. G, S.~D, T.~Keerthi, R.~S. R, \href{http://dx.doi.org/10.1109/ICACITE53722.2022.9823806}{Mnist handwritten digit recognition using machine learning}, in: 2022 2nd International Conference on Advance Computing and Innovative Technologies in Engineering (ICACITE), IEEE, 2022, p. 768–772.
\newblock \href {https://doi.org/10.1109/icacite53722.2022.9823806} {\path{doi:10.1109/icacite53722.2022.9823806}}.
\newline\urlprefix\url{http://dx.doi.org/10.1109/ICACITE53722.2022.9823806}

\bibitem{lda}
M.~Lasalvia, V.~Capozzi, G.~Perna, \href{http://dx.doi.org/10.3390/app12115345}{A comparison of pca-lda and pls-da techniques for classification of vibrational spectra}, Applied Sciences 12~(11) (2022) 5345.
\newblock \href {https://doi.org/10.3390/app12115345} {\path{doi:10.3390/app12115345}}.
\newline\urlprefix\url{http://dx.doi.org/10.3390/app12115345}

\bibitem{iris_folower_dataset}
S.~Goyal, A.~Sharma, P.~Gupta, P.~Chandi, \href{http://dx.doi.org/10.1007/978-981-16-1048-6_50}{Assessment of Iris Flower Classification Using Machine Learning Algorithms}, Springer Singapore, 2021, p. 641–649.
\newblock \href {https://doi.org/10.1007/978-981-16-1048-6_50} {\path{doi:10.1007/978-981-16-1048-6_50}}.
\newline\urlprefix\url{http://dx.doi.org/10.1007/978-981-16-1048-6_50}

\bibitem{r1_svm}
S.~O. Oladejo, S.~O. Ekwe, A.~T. Ajibare, L.~A. Akinyemi, S.~Mirjalili, \href{http://dx.doi.org/10.1016/B978-0-32-395365-8.00042-7}{Tuning SVMs’ hyperparameters using the whale optimization algorithm}, Elsevier, 2024, p. 495–521.
\newblock \href {https://doi.org/10.1016/b978-0-32-395365-8.00042-7} {\path{doi:10.1016/b978-0-32-395365-8.00042-7}}.
\newline\urlprefix\url{http://dx.doi.org/10.1016/B978-0-32-395365-8.00042-7}

\bibitem{r3_improved_mlp}
M.~Kumar, U.~Mehta, G.~Cirrincione, \href{http://dx.doi.org/10.1016/j.aiopen.2023.12.003}{Enhancing neural network classification using fractional-order activation functions}, AI Open 5 (2024) 10–22.
\newblock \href {https://doi.org/10.1016/j.aiopen.2023.12.003} {\path{doi:10.1016/j.aiopen.2023.12.003}}.
\newline\urlprefix\url{http://dx.doi.org/10.1016/j.aiopen.2023.12.003}

\bibitem{breast_cancer_dataset}
M.~H. Alshayeji, H.~Ellethy, S.~Abed, R.~Gupta, \href{http://dx.doi.org/10.1016/j.bspc.2021.103141}{Computer-aided detection of breast cancer on the wisconsin dataset: An artificial neural networks approach}, Biomedical Signal Processing and Control 71 (2022) 103141.
\newblock \href {https://doi.org/10.1016/j.bspc.2021.103141} {\path{doi:10.1016/j.bspc.2021.103141}}.
\newline\urlprefix\url{http://dx.doi.org/10.1016/j.bspc.2021.103141}

\bibitem{uot_3_woa}
Z.~Waheed, A.~Humaidi, \href{https://ijccce.uotechnology.edu.iq/article\_178337.html}{Whale optimization algorithm enhances the performance of knee-exoskeleton system controlled by smc}, IRAQI JOURNAL OF COMPUTERS, COMMUNICATIONS, CONTROL AND SYSTEMS ENGINEERING 23~(2) (2023) 125--135.
\newblock \href {https://doi.org/https://doi.org/10.33103/uot.ijccce.23.2.10} {\path{doi:https://doi.org/10.33103/uot.ijccce.23.2.10}}.
\newline\urlprefix\url{https://ijccce.uotechnology.edu.iq/article\_178337.html}

\bibitem{num_liver_lobules}
N.~J. Krebs, C.~Neville, J.~P. Vacanti, \href{http://dx.doi.org/10.1016/B978-012369415-7/50013-2}{Cellular Transplants for Liver Diseases}, Elsevier, 2007, p. 215–240.
\newblock \href {https://doi.org/10.1016/b978-012369415-7/50013-2} {\path{doi:10.1016/b978-012369415-7/50013-2}}.
\newline\urlprefix\url{http://dx.doi.org/10.1016/B978-012369415-7/50013-2}

\bibitem{softmax}
J.~S. Bridle, \href{http://dx.doi.org/10.1007/978-3-642-76153-9_28}{Probabilistic Interpretation of Feedforward Classification Network Outputs, with Relationships to Statistical Pattern Recognition}, Springer Berlin Heidelberg, 1990, p. 227–236.
\newblock \href {https://doi.org/10.1007/978-3-642-76153-9_28} {\path{doi:10.1007/978-3-642-76153-9_28}}.
\newline\urlprefix\url{http://dx.doi.org/10.1007/978-3-642-76153-9_28}

\bibitem{softmax_1_tarik}
A.~Arora, O.~H. Alsadoon, T.~W.~A. Khairi, T.~A. Rashid, \href{http://dx.doi.org/10.1109/CITISIA50690.2020.9371821}{A novel softmax regression enhancement for handwritten digits recognition using tensor flow library}, in: 2020 5th International Conference on Innovative Technologies in Intelligent Systems and Industrial Applications (CITISIA), IEEE, 2020, p. 1–9.
\newblock \href {https://doi.org/10.1109/citisia50690.2020.9371821} {\path{doi:10.1109/citisia50690.2020.9371821}}.
\newline\urlprefix\url{http://dx.doi.org/10.1109/CITISIA50690.2020.9371821}

\bibitem{softmax_2}
S.~Maharjan, A.~Alsadoon, P.~Prasad, T.~Al-Dalain, O.~H. Alsadoon, \href{http://dx.doi.org/10.1016/j.jneumeth.2019.108520}{A novel enhanced softmax loss function for brain tumour detection using deep learning}, Journal of Neuroscience Methods 330 (2020) 108520.
\newblock \href {https://doi.org/10.1016/j.jneumeth.2019.108520} {\path{doi:10.1016/j.jneumeth.2019.108520}}.
\newline\urlprefix\url{http://dx.doi.org/10.1016/j.jneumeth.2019.108520}

\bibitem{fox}
H.~Mohammed, T.~Rashid, \href{http://dx.doi.org/10.1007/s10489-022-03533-0}{Fox: a fox-inspired optimization algorithm}, Applied Intelligence 53~(1) (2022) 1030–1050.
\newblock \href {https://doi.org/10.1007/s10489-022-03533-0} {\path{doi:10.1007/s10489-022-03533-0}}.
\newline\urlprefix\url{http://dx.doi.org/10.1007/s10489-022-03533-0}

\bibitem{fox_tsa}
S.~A. Aula, T.~A. Rashid, \href{http://dx.doi.org/10.1016/j.asej.2024.103185}{Fox-tsa: Navigating complex search spaces and superior performance in benchmark and real-world optimization problems}, Ain Shams Engineering Journal 16~(1) (2025) 103185.
\newblock \href {https://doi.org/10.1016/j.asej.2024.103185} {\path{doi:10.1016/j.asej.2024.103185}}.
\newline\urlprefix\url{http://dx.doi.org/10.1016/j.asej.2024.103185}

\bibitem{epsilon_greedy}
Y.~Liu, B.~Cao, H.~Li, \href{http://dx.doi.org/10.1007/s40747-020-00138-3}{Improving ant colony optimization algorithm with epsilon greedy and levy flight}, Complex \&; Intelligent Systems 7~(4) (2020) 1711–1722.
\newblock \href {https://doi.org/10.1007/s40747-020-00138-3} {\path{doi:10.1007/s40747-020-00138-3}}.
\newline\urlprefix\url{http://dx.doi.org/10.1007/s40747-020-00138-3}

\bibitem{uot_alaa}
Z.~K. Abdalrdha, A.~M. Al-Bakry, A.~K. Farhan, \href{http://dx.doi.org/10.1109/DeSE60595.2023.10469361}{Cnn hyper-parameter optimizer based on evolutionary selection and gow approach for crimes tweet detection}, in: 2023 16th International Conference on Developments in eSystems Engineering (DeSE), IEEE, 2023.
\newblock \href {https://doi.org/10.1109/dese60595.2023.10469361} {\path{doi:10.1109/dese60595.2023.10469361}}.
\newline\urlprefix\url{http://dx.doi.org/10.1109/DeSE60595.2023.10469361}

\bibitem{log_loss}
Y.~XUE, G.~JIN, T.~SHEN, L.~TAN, L.~WANG, \href{http://dx.doi.org/10.1016/j.cja.2023.03.048}{Template-guided frequency attention and adaptive cross-entropy loss for uav visual tracking}, Chinese Journal of Aeronautics 36~(9) (2023) 299–312.
\newblock \href {https://doi.org/10.1016/j.cja.2023.03.048} {\path{doi:10.1016/j.cja.2023.03.048}}.
\newline\urlprefix\url{http://dx.doi.org/10.1016/j.cja.2023.03.048}

\bibitem{evaluation_metrics}
G.~Naidu, T.~Zuva, E.~M. Sibanda, \href{http://dx.doi.org/10.1007/978-3-031-35314-7_2}{A Review of Evaluation Metrics in Machine Learning Algorithms}, Springer International Publishing, 2023, p. 15–25.
\newblock \href {https://doi.org/10.1007/978-3-031-35314-7_2} {\path{doi:10.1007/978-3-031-35314-7_2}}.
\newline\urlprefix\url{http://dx.doi.org/10.1007/978-3-031-35314-7_2}

\bibitem{goose}
R.~K. Hamad, T.~A. Rashid, \href{http://dx.doi.org/10.1007/s12530-023-09553-6}{Goose algorithm: a powerful optimization tool for real-world engineering challenges and beyond}, Evolving Systems 15~(4) (2024) 1249–1274.
\newblock \href {https://doi.org/10.1007/s12530-023-09553-6} {\path{doi:10.1007/s12530-023-09553-6}}.
\newline\urlprefix\url{http://dx.doi.org/10.1007/s12530-023-09553-6}

\bibitem{ana}
D.~N. Hama~Rashid, T.~A. Rashid, S.~Mirjalili, \href{https://www.mdpi.com/2227-7390/9/23/3111}{Ana: Ant nesting algorithm for optimizing real-world problems}, Mathematics 9~(23) (2021).
\newblock \href {https://doi.org/10.3390/math9233111} {\path{doi:10.3390/math9233111}}.
\newline\urlprefix\url{https://www.mdpi.com/2227-7390/9/23/3111}

\bibitem{leo}
A.~M. Aladdin, T.~A. Rashid, \href{http://dx.doi.org/10.1007/s00521-025-11225-2}{Leo: Lagrange elementary optimization}, Neural Computing and Applications 37~(19) (2025) 14365–14397.
\newblock \href {https://doi.org/10.1007/s00521-025-11225-2} {\path{doi:10.1007/s00521-025-11225-2}}.
\newline\urlprefix\url{http://dx.doi.org/10.1007/s00521-025-11225-2}

\bibitem{fdo}
J.~M. Abdullah, T.~Ahmed, \href{http://dx.doi.org/10.1109/ACCESS.2019.2907012}{Fitness dependent optimizer: Inspired by the bee swarming reproductive process}, IEEE Access 7 (2019) 43473–43486.
\newblock \href {https://doi.org/10.1109/access.2019.2907012} {\path{doi:10.1109/access.2019.2907012}}.
\newline\urlprefix\url{http://dx.doi.org/10.1109/ACCESS.2019.2907012}

\bibitem{wilcoxon_signed_rank_test}
C.~B. Hodges, B.~M. Stone, P.~K. Johnson, J.~H. Carter, C.~K. Sawyers, P.~R. Roby, H.~M. Lindsey, \href{http://dx.doi.org/10.3758/s13428-022-01932-2}{Researcher degrees of freedom in statistical software contribute to unreliable results: A comparison of nonparametric analyses conducted in spss, sas, stata, and r}, Behavior Research Methods 55~(6) (2022) 2813–2837.
\newblock \href {https://doi.org/10.3758/s13428-022-01932-2} {\path{doi:10.3758/s13428-022-01932-2}}.
\newline\urlprefix\url{http://dx.doi.org/10.3758/s13428-022-01932-2}

\bibitem{accuracy_Confidence}
Y.~Qu, K.~Roitero, D.~L. Barbera, D.~Spina, S.~Mizzaro, G.~Demartini, \href{http://dx.doi.org/10.1145/3546916}{Combining human and machine confidence in truthfulness assessment}, Journal of Data and Information Quality 15~(1) (2022) 1–17.
\newblock \href {https://doi.org/10.1145/3546916} {\path{doi:10.1145/3546916}}.
\newline\urlprefix\url{http://dx.doi.org/10.1145/3546916}

\bibitem{sdg}
S.~Wojtowytsch, \href{http://dx.doi.org/10.1007/s00332-023-09903-3}{Stochastic gradient descent with noise of machine learning type part i: Discrete time analysis}, Journal of Nonlinear Science 33~(3) (Mar. 2023).
\newblock \href {https://doi.org/10.1007/s00332-023-09903-3} {\path{doi:10.1007/s00332-023-09903-3}}.
\newline\urlprefix\url{http://dx.doi.org/10.1007/s00332-023-09903-3}

\end{thebibliography}
\end{document}